\documentclass{article}

\usepackage{iclr2027_conference,times}
\usepackage[utf8]{inputenc}
\usepackage[T1]{fontenc}
\usepackage[hidelinks]{hyperref}
\usepackage{url}
\usepackage{booktabs}
\usepackage{amsmath}
\usepackage{amssymb}
\usepackage{graphicx}
\usepackage{microtype}
\usepackage{xcolor}
\usepackage{enumitem}
\usepackage{multirow}
\usepackage{wrapfig}
\usepackage{placeins}

\newcommand{\method}{G$^3$-LoRA}
\newcommand{\grad}{\nabla}
\newcommand{\loss}{\mathcal{L}}
\newcommand{\data}{\mathcal{D}}
\newcommand{\sg}{\operatorname{sg}}

\iclrfinalcopy

\title{G$^3$-LoRA: Organizing Reward-Weighted Video Data with \textbf{G}radient-\textbf{G}uided \textbf{G}rouped LoRA}

\author{%
  Jia Song$^{1,*}$ \quad Wenhow Li$^{1,*}$ \quad Lichen Bai$^{1}$ \quad Bada Ye$^{2}$ \quad Zeke Xie$^{1,\dagger}$\\[4pt]
  \normalfont $^{1}$The Hong Kong University of Science and Technology (Guangzhou)\\
  \normalfont $^{2}$Tencent\\[3pt]
  \normalfont\small $^{*}$Equal contribution. \quad $^{\dagger}$Corresponding author.\\
  \normalfont\small\texttt{jsong231@connect.hkust-gz.edu.cn} \quad $^{\dagger}$\texttt{zekexie@hkust-gz.edu.cn}
}

\hypersetup{
  pdftitle={G3-LoRA: Organizing Reward-Weighted Video Data with Gradient-Guided Grouped LoRA},
  pdfauthor={Jia Song, Wenhow Li, Lichen Bai, Bada Ye, Zeke Xie}
}

\begin{document}
\maketitle

\begin{abstract}
Post-training foundation video models on heterogeneous reward-weighted data usually assumes that all data categories induce compatible updates. This assumption is fragile when categories correspond to different skills, domains, or evaluation dimensions. We study this problem in text-to-video post-training, where VBench2.0 dimensions define data buckets and an external multimodal reward pipeline assigns sample weights. We propose \method{} (\textbf{G}radient-\textbf{G}uided \textbf{G}rouped LoRA), a data organization procedure that probes category-level gradients induced by reward-weighted video samples, removes the shared global update direction, clusters categories by residual gradient compatibility, trains group-specific LoRA experts, and consolidates them into one adapter by weight merging followed by on-policy distillation from the experts. We motivate this procedure by viewing reward-weighted flow matching as velocity-field regression: incompatible reward dimensions may prefer different denoising directions in overlapping noisy latent regions, causing shared LoRA training to average capabilities. On Wan2.1-T2V-1.3B-Diffusers, the merged grouped adapter improves the matched VBench2.0 evaluation over the base model, a joint reward-weighted LoRA baseline, and random, semantic, and raw-gradient partitions trained with the same pipeline; an independent evaluator agrees, and on CogVideoX-2B grouping avoids the negative transfer of joint training. The gain is not uniform: merging compresses the largest specialist gains, distillation recovers part of this loss, and camera motion and several local-quality dimensions remain challenging. Together, these results suggest that gradient compatibility can serve as a practical diagnostic for organizing reward-weighted video post-training data.
\end{abstract}

\section{Introduction}

Post-training pipelines increasingly rely on heterogeneous supervision. A single adapter may be trained on data from multiple domains, tasks, preference sources, or quality dimensions. The default engineering choice is simple: mix all data and optimize one model. This choice is convenient, but it hides a strong assumption: data categories that share a model should also share compatible optimization directions.

This paper asks whether that assumption should be measured rather than accepted. We focus on text-to-video generation because evaluation already decomposes model quality into multiple dimensions. VBench and VBench2.0 evaluate properties such as human identity, camera motion, material change, multi-view consistency, and complex plot faithfulness~\citep{huang2023vbench,zheng2025vbench2}. These dimensions are usually treated as reporting axes after generation, but they can also define post-training data buckets. A model can generate candidate videos for each bucket, an external multimodal model can score them, and the resulting reward-weighted data can be used to train a LoRA adapter~\citep{hu2021lora}. The open question is how those buckets should be combined before adapter training.

The answer is not obvious. Some dimensions appear semantically related but may prefer different local updates; others appear unrelated but may share geometric, temporal, or object-centric structure. Human identity consistency may reward stable subject appearance, while dynamic attribute tasks emphasize visible change. Camera motion and multi-view consistency may share 3D structure, but they can differ in how much motion the generated video should contain. A single mixed adapter may improve average quality while quietly degrading conflict-heavy dimensions. Conversely, training one adapter per dimension avoids interference but loses data sharing. We therefore treat category organization as a first-class post-training problem.

The key mechanism is specific to diffusion and flow SFT. Sample-level rewards do not directly optimize a scalar preference at inference time; instead, they reweight the regression of the denoising or velocity field. Each evaluation dimension can therefore be viewed as inducing a reward-reweighted conditional velocity field. When multiple dimensions favor incompatible generation behaviors, their preferred velocity directions may conflict in overlapping noisy latent regions. Since the MSE objective learns a conditional mean target, joint training with a shared LoRA can average these directions, producing gradient interference and capability averaging. To address this, we organize data categories by their measured optimization compatibility before training grouped adapters.

Therefore, we propose \method{}: \textbf{G}radient-\textbf{G}uided \textbf{G}rouped LoRA. The method first computes a reward-weighted gradient for each data category using the same flow-matching loss used in training~\citep{lipman2022flow}. It then computes a global mixed gradient and removes the component of each category gradient explained by this shared direction. The residual gradients define a compatibility matrix: positive cosine similarity suggests that two categories induce similar category-specific update directions, while negative similarity suggests potential interference. We cluster this matrix into groups, train one LoRA expert per group, and merge the experts into a final adapter. Because merging can dilute group-specific updates, we then distill the experts into the merged adapter on its own sampling trajectories (on-policy distillation, OPD)~\citep{agarwal2024onpolicy,fang2026flowopd}.

The concrete instantiation uses Wan2.1-T2V-1.3B-Diffusers~\citep{wan2025,wan2025huggingface,diffusers2024}, 17 non-diversity VBench2.0 dimensions, and Qwen3-VL-8B-Instruct-based QA rewards~\citep{bai2025qwen3vl}; we repeat the procedure on CogVideoX-2B~\citep{yang2025cogvideox}, which differs in architecture and training objective. The broader point is not specific to this benchmark. VBench2.0 provides a convenient taxonomy and evaluator, but the proposed procedure applies whenever a post-training dataset can be partitioned into reward-bearing categories whose interactions are uncertain.

\paragraph{Contributions.}
\begin{itemize}[leftmargin=*, itemsep=1pt, topsep=2pt]
    \item We formulate heterogeneous video post-training as a data organization problem: categories should be grouped according to optimization compatibility, not only semantic labels.
    \item We connect reward-weighted flow SFT to reward-reweighted velocity-field regression, providing a local mechanism for capability averaging under shared LoRA training.
    \item We introduce a residual gradient compatibility measure that subtracts the global mixed update direction before comparing category-specific gradients, and show that the conflict it reveals is reproducible across probe sets and backbones and predicts held-out transfer.
    \item We instantiate the method on two video backbones with matched grouping controls, an independent evaluator, and paired bootstrap intervals, and we locate where merging loses specialist capability and how much on-policy distillation recovers.
\end{itemize}

\section{Method}

\subsection{Problem setting}

Let $\data=\{\data_1,\ldots,\data_K\}$ be a post-training dataset partitioned into $K$ categories. In our implementation, $K=17$ and each category corresponds to a non-diversity VBench2.0 dimension. Training videos are generated by the same base model that will later be post-trained, so the dataset is a self-generated improvement pool rather than a separately collected human-video corpus. A sample $x=(v,p,c,w)$ contains a generated video $v$, prompt $p$, category $c$, and scalar reward weight $w$. The reward comes from a dimension-specific QA pipeline: a multimodal model observes the video, answers generated questions, and the fraction of expected answers is converted into a sample-level training weight. The weight is used as a scalar multiplier on the flow-matching loss; it is not an RL objective and does not require preference-pair optimization.

We train only LoRA parameters $\phi$ on top of a frozen video generation model. For each video, the VAE encodes frames into a latent $z_1$, noise $z_0\sim\mathcal{N}(0,I)$ is sampled, and a time $t$ defines the flow-matching interpolation as
\begin{equation}
    z_t=(1-t)z_0+t z_1,\qquad u=z_1-z_0.
    \label{eq:flow_interpolation}
\end{equation}
The model predicts $\hat{u}_\phi(z_t,p,t)$ and optimizes a reward-weighted loss as
\begin{equation}
    \loss(\phi; x)=\frac{w}{\bar{w}}\left\|\hat{u}_\phi(z_t,p,t)-u\right\|_2^2,
    \label{eq:reward_loss}
\end{equation}
where $\bar{w}$ is the global mean reward weight. Reward weighting is not only sample filtering: it changes which videos dominate the category gradient and therefore changes the compatibility structure that \method{} estimates. The naive joint baseline minimizes the average of Eq.~\ref{eq:reward_loss} across all categories. \method{} instead uses gradients of Eq.~\ref{eq:reward_loss} to decide which categories should share an adapter.

\paragraph{Capability averaging under MSE regression.}
For a fixed prompt and nearby noisy latent region, suppose two reward dimensions emphasize different target velocities $u_1$ and $u_2$. A shared MSE regressor trained on both distributions prefers a conditional mean direction rather than either specialized direction. In video generation, this can manifest as softened camera motion, weaker attribute change, or partial preservation of structure without fully satisfying the corresponding dimension. We use this mechanism as the local explanation for why reward-weighted post-training can improve the mean score while still degrading conflict-heavy dimensions.

\subsection{Gradient-guided grouped LoRA}

\begin{figure}[t]
    \centering
    \includegraphics[width=\linewidth]{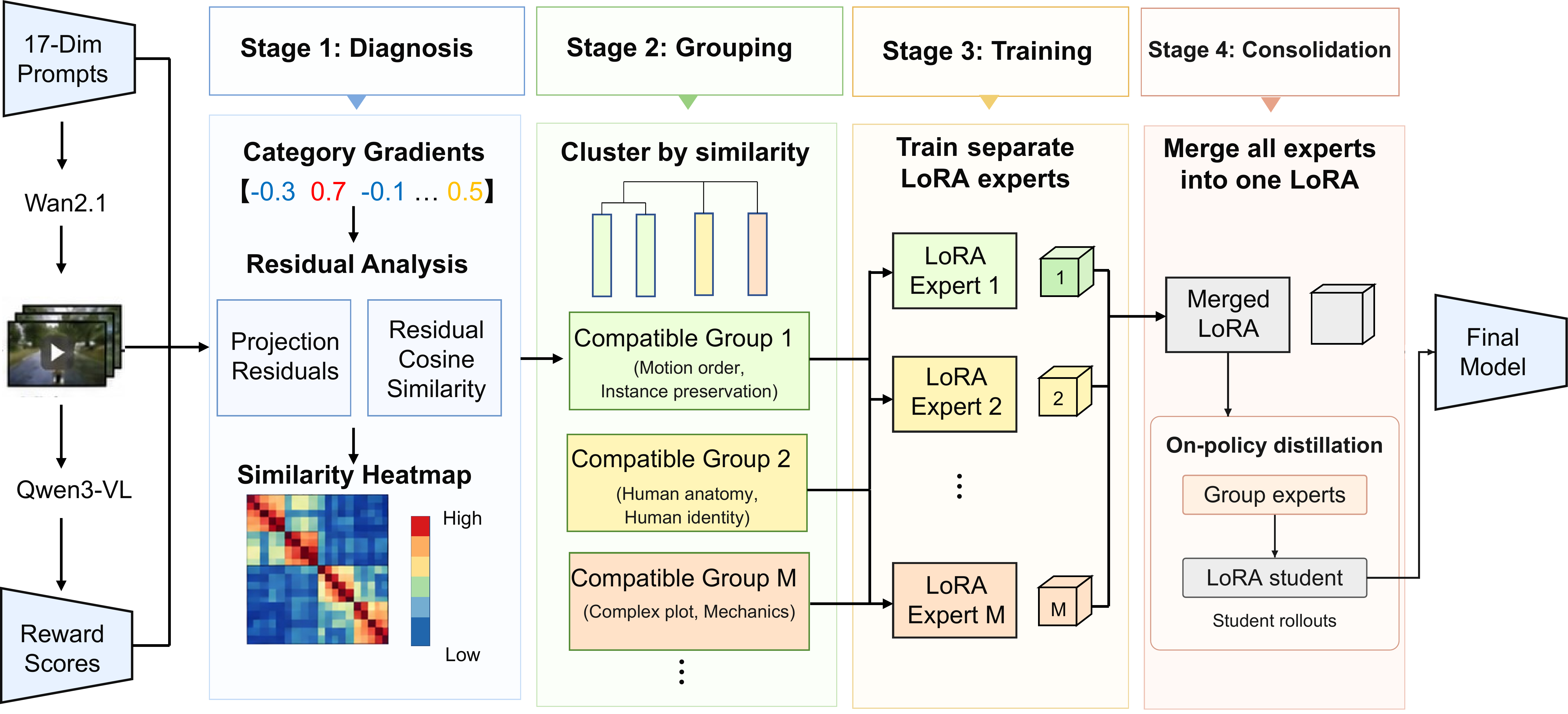}
    \caption{\textbf{Method overview.} \method{} treats post-training categories as optimization objects. It uses reward-weighted category gradients to estimate compatibility, clusters categories after removing the global mixed direction, trains group-specific LoRA experts, merges them, and distills the experts back into the merged adapter.}
    \label{fig:method}
\end{figure}

\paragraph{Reward-weighted category probing.}
For each category $i$, we estimate a reward-weighted LoRA gradient as
\begin{equation}
    g_i = \frac{1}{|\mathcal{B}_i|}\sum_{x\in \mathcal{B}_i}\grad_\phi \loss(\phi;x),
    \label{eq:category_gradient}
\end{equation}
where $\mathcal{B}_i$ is a fixed probe subset. To reduce probe noise, the implementation fixes diffusion timesteps and noise seeds during probing. Probing is repeated at multiple training stages, e.g., 0\%, 25\%, 50\%, 75\%, and 100\% of a reference joint training trajectory. The resulting gradient is a tractable proxy for the reward-reweighted velocity field induced by a category.

\paragraph{Removing the shared update direction.}
Raw category gradients can be dominated by a shared update that is useful for all categories, such as adapting the model to the video resolution, prompt distribution, or LoRA parameterization. Comparing raw gradients can therefore overstate compatibility because the common direction can mask category-specific conflicts. We compute a global gradient $g_{\mathrm{global}}$ on a mixed probe batch and remove its projection from each category gradient as
\begin{equation}
    r_i = g_i - \frac{\langle g_i, g_{\mathrm{global}}\rangle}{\|g_{\mathrm{global}}\|_2^2+\epsilon}g_{\mathrm{global}}.
    \label{eq:residual}
\end{equation}
The residual $r_i$ captures category-specific pressure after accounting for the common direction. Directly subtracting $g_{\mathrm{global}}$ instead gives the same partition (adjusted Rand index 1.0; Appendix~\ref{app:gradient_diagnostics}), so the essential operation is removing the shared direction rather than the specific projection operator.

\paragraph{Compatibility and grouping.}
The residual compatibility between categories $i$ and $j$ is defined as
\begin{equation}
    S_{ij} = \frac{\langle r_i,r_j\rangle}{\|r_i\|_2\|r_j\|_2+\epsilon}.
    \label{eq:residual_similarity}
\end{equation}
When multiple probe stages are available, we average $S_{ij}$ across stages to obtain $\bar{S}_{ij}$. We then run average-linkage hierarchical clustering with distance $D_{ij}=1-\bar{S}_{ij}$ and select $M$ groups. The experiments use $M=5$.

\paragraph{Grouped LoRA training and merge.}
For each group $G_m$, we train an independent LoRA expert using only samples from categories in $G_m$ and the reward-weighted objective in Eq.~\ref{eq:reward_loss}. Group-specific training prevents strongly conflicting categories from sharing all update steps, while still allowing compatible categories to share data. The final adapter is obtained by merging group experts into a single deployable LoRA module with a dense delta-SVD procedure: group LoRA updates are expanded to dense weight deltas, averaged with group-size weights, and recompressed into one LoRA adapter. Appendix~\ref{app:merge} gives the dense merge and recomposition equations. This can be viewed as a weight-space approximation to an ensemble or mixture of skill-specific experts: training creates several local specialists, while merging folds them into one adapter for inference.

\paragraph{Merge retention.}
The success of grouped training depends on whether the learned LoRA updates remain compatible after merging. If two group deltas satisfy $\Delta W_i\approx-\Delta W_j$, uniform averaging can cancel both capabilities. Even when parameter-space cosine is small, the updates may still conflict on real activations. We therefore treat merge retention as a diagnostic: a merged adapter should preserve a substantial fraction of each group expert's gain on the dimensions inside that group.

\paragraph{On-policy expert distillation (OPD).}
When merging loses part of an expert's gain, we fine-tune the merged adapter $\phi_0$ so that, for each prompt, its velocity matches the expert $\psi_{m(p)}$ of the prompt's group on latent states $z_\tau$ drawn from the student's own sampling trajectories $\rho_\phi(\cdot\mid p)$. Initializing $\phi\leftarrow\phi_0$, OPD minimizes
\begin{equation}
\begin{split}
    \loss_{\mathrm{OPD}}(\phi)=\mathbb{E}_{p,\;z_\tau\sim\rho_\phi(\cdot\mid p)}\Big[&\big\|\hat{u}_\phi(z_\tau,p,\tau)-\sg\,\hat{u}_{\psi_{m(p)}}(z_\tau,p,\tau)\big\|_2^2\\
    &+\lambda\big\|\hat{u}_\phi(z_\tau,p,\tau)-\sg\,\hat{u}_{\phi_0}(z_\tau,p,\tau)\big\|_2^2\Big],
\end{split}
    \label{eq:opd}
\end{equation}
where $\sg$ is stop-gradient and the second term anchors the student to the merged adapter. Sampling states from the student, rather than noising training videos, supervises it on the latents it visits at inference~\citep{agarwal2024onpolicy,fang2026flowopd}. The loss uses neither reward weights nor training videos (Appendix~\ref{app:opd}).

\section{Experiments}

\subsection{Experimental setup}

\paragraph{Model and data.}
The base model is Wan2.1-T2V-1.3B-Diffusers. Training videos are generated by the same base model from dimension-specific prompts. Each of 17 non-diversity VBench2.0 dimensions contributes approximately 560 videos. Qwen3-VL-8B-Instruct provides reward weights through the dimension-specific QA pipeline; Appendix~\ref{app:reward_pipeline} describes this reward construction. As a second backbone and objective, we rerun the full procedure on CogVideoX-2B with its native diffusion-denoising loss and rank-128 experts merged exactly into a rank-640 adapter (Appendix~\ref{app:cogvideox}).

\paragraph{Evaluation.}
We evaluate on VBench2.0 dimensions excluding diversity. All main comparisons use a matched evaluation protocol with aligned prompts, seeds, generation parameters, and evaluator settings; Appendix~\ref{app:training_eval_params} gives the concrete generation and evaluation details. Uncertainty is estimated by a paired bootstrap over prompts (Appendix~\ref{app:videoscore}). As an evaluator independent of the Qwen reward model, we also score the same videos with VideoScore-v1.1~\citep{he2024videoscore}. The retention study in Appendix~\ref{app:retention} is a separate matched diagnostic evaluation and is reported only for analyzing merge behavior.

\paragraph{Baselines and compared adapters.}
The main comparison includes Base Wan2.1 without post-training; Joint LoRA trained on all 17 reward-weighted dimensions; four $M=5$ partition controls, namely two random partitions, the official VBench2.0 semantic categories, and clustering of raw (unprojected) gradients; the merged Grouped LoRA produced by the dense delta-SVD-style merge; and Grouped LoRA + OPD. The partition controls use the same grouped-LoRA training and merge pipeline as \method{}, but their dimension partitions are not selected by residual-gradient compatibility; Appendix~\ref{app:control_grouping} lists all partitions. A separate retention evaluation compares the base model, the merged grouped adapter, and each group-only LoRA evaluated only on the VBench2.0 dimensions assigned to that group.
\subsection{Gradient grouping analysis}

The gradient analysis produces five groups over the 17 non-diversity VBench2.0 dimensions:

\begin{itemize}[leftmargin=*, itemsep=1pt, topsep=2pt]
    \item Group 1: dynamic spatial relationship, human clothes consistency, instance preservation, motion order.
    \item Group 2: human anatomy, human identity consistency, human interaction, motion rationality.
    \item Group 3: camera motion, complex landscape, composition.
    \item Group 4: dynamic attribute, material, multi-view consistency, thermotics.
    \item Group 5: complex plot, mechanics.
\end{itemize}

\begin{figure}[t]
    \centering
    \includegraphics[width=0.67\linewidth]{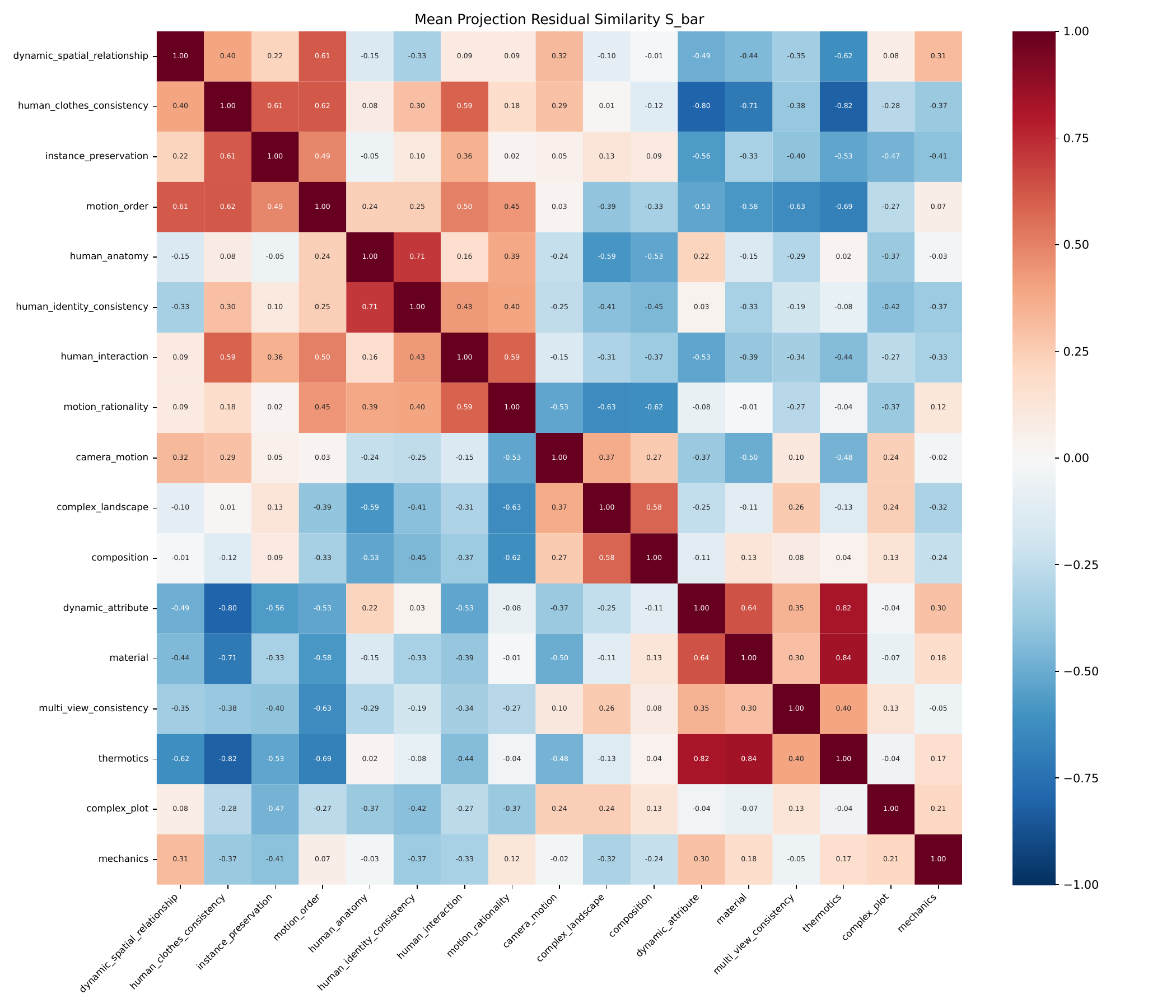}
    \caption{\textbf{Residual gradient compatibility across VBench2.0 dimensions.} Each entry is the mean cosine similarity between two category gradients after projecting out the global mixed-gradient direction. Dimensions are ordered by the final $M=5$ grouping, so block structure indicates compatible categories and blue off-block regions indicate potential interference.}
    \label{fig:gradient_heatmap}
\end{figure}

Figure~\ref{fig:gradient_heatmap} visualizes the matrix used for grouping. The strongest positive pairs are not arbitrary semantic neighbors: material and thermotics have mean similarity 0.841, dynamic attribute and thermotics have 0.824, and human anatomy and human identity consistency have 0.709. Conversely, the stable negative pairs reveal conflicts that a naive semantic grouping could miss, such as human clothes consistency versus thermotics (-0.818), dynamic attribute versus human clothes consistency (-0.798), and motion order versus multi-view consistency (-0.630). These values are averaged over five probe stages, making the matrix a staged compatibility estimate rather than a single-batch diagnostic.

We choose $M=5$ because it gives a cleaner residual-gradient partition than $M=4$. Under average-linkage clustering on $D=1-\bar{S}$, the $M=5$ split increases within-group mean similarity from 0.345 to 0.473, improves the within-minus-between separation from 0.559 to 0.633, and removes all negative within-group pairs. In contrast, the $M=4$ split leaves three negative within-group pairs, including one strongly negative pair. The grouping is therefore selected to reduce within-adapter gradient conflict rather than to match a hand-written semantic taxonomy.

This grouping has two uses in the paper. First, it is a diagnostic result: it shows that reward-weighted category gradients contain structure that is not visible from metric names alone. Second, it defines the grouped-LoRA training plan: each group trains one LoRA expert, and the experts are merged before VBench2.0 evaluation. The downstream results below show that the grouping improves the average score and reduces the worst dimension-level drop, but also exposes a dimension-level tradeoff: camera motion does not inherit the improvement obtained by joint LoRA. Appendix~\ref{app:gradient_diagnostics} provides the corresponding dendrogram and stage-wise heatmaps.

This conflict is hidden by the shared direction and is not specific to one backbone. With raw gradients, all 136 category pairs on Wan2.1 have positive cosine similarity; after removal, 75 pairs (55\%) are negative, and on CogVideoX-2B 81--87 pairs are negative across three non-overlapping probe cohorts. Resampled probe sets reproduce the Wan2.1 matrix (Pearson 0.847), and residual similarity predicts how a step along one category's gradient changes another category's held-out loss (Spearman $\rho=0.256$, permutation $p=0.003$).

\subsection{VBench2.0 results}

\begin{table}[t]
    \centering
    \caption{\textbf{VBench2.0 comparison on Wan2.1.} Grouped LoRA (Ours) obtains the best mean among merged adapters and a smaller worst drop than joint LoRA or the random partitions; OPD raises the mean further. Ties count as neither improved nor degraded; ``--'': not computed.}
    \label{tab:main_results}
    \small
    \begin{tabular}{lrrrr}
        \toprule
        Method & Mean & $\Delta$ Mean & Improved / degraded dims & Worst drop vs. Base \\
        \midrule
        Base Wan2.1 & 47.07 & -- & -- & -- \\
        Joint LoRA & 46.99 & -0.07 & 9 / 8 & -8.00 \\
        Random partition A & 46.38 & -0.69 & 6 / 9 & -8.21 \\
        Random partition B & 45.82 & -1.25 & 8 / 8 & -16.00 \\
        Semantic partition & 47.20 & +0.13 & -- & -- \\
        Raw-gradient partition & 47.35 & +0.28 & -- & -- \\
        Grouped LoRA (Ours) & 48.18 & +1.11 & 9 / 6 & -4.52 \\
        Grouped LoRA + OPD (Ours) & \textbf{49.91} & \textbf{+2.84} & \textbf{12 / 4} & \textbf{-3.97} \\
        \bottomrule
    \end{tabular}
\end{table}

Table~\ref{tab:main_results} gives the downstream picture. Joint LoRA is slightly below the base model in mean score (-0.07), despite producing gains on camera motion (+8.00), mechanics (+5.18), multi-view consistency (+4.64), and dynamic spatial relationship (+4.00). This pattern is consistent with a single mixed adapter under heterogeneous reward weighting: some dimensions benefit, but the adapter also pays for the mixture with regressions in human interaction (-8.00), material (-7.63), and dynamic attribute (-6.00).

The two random partitions are also weak: Random partition A reaches 46.38 mean score, while Random partition B reaches 45.82. Both are below the base model, joint LoRA, and Grouped LoRA. Random partition A improves 6 out of 17 dimensions with a worst dimension-level drop of -8.21, and Random partition B improves 8 dimensions but has a larger worst drop of -16.00. The semantic and raw-gradient partitions are only marginally above the base model (47.20 and 47.35); the semantic partition places 13 negatively related pairs in the same group, whereas the residual partition places none. These comparisons indicate that the benefit of the grouped adapter is not explained by simply splitting the 17 dimensions into several smaller LoRA experts; the partition itself matters.

Grouped LoRA increases the mean to 48.18, a +1.11 gain over the base model and a +1.19 gain over joint LoRA. The largest gains over the base model occur on multi-view consistency (+8.99), motion order understanding (+8.00), dynamic spatial relationship (+4.00), thermotics (+3.29), and complex landscape (+2.40). These improvements support the main hypothesis that grouping can reduce some forms of capability averaging: the grouped adapter improves the average tradeoff and substantially reduces the worst dimension-level drop compared with joint LoRA. The gain over the base model is positive in 96.6\% of paired bootstrap resamples (95\% CI $[-0.08, +2.26]$), so it is a consistent trend rather than a significant difference at the 5\% level.

At the same time, the result is not a clean dominance claim. The grouped adapter still degrades material (-4.52), human interaction (-4.00), and several identity or clothing dimensions relative to the base model. It also gives up the joint LoRA's camera-motion gain: camera motion remains at the base score of 12.00, while joint LoRA reaches 20.00. This is the clearest diagnostic contrast in Appendix Table~\ref{tab:full_scores}. VBench2.0 camera motion is a hard label match over CoTracker-estimated motion classes, so improving the mean score and multi-view consistency does not guarantee better camera-trajectory control.

\begin{figure}[t]
    \centering
    \includegraphics[width=\linewidth]{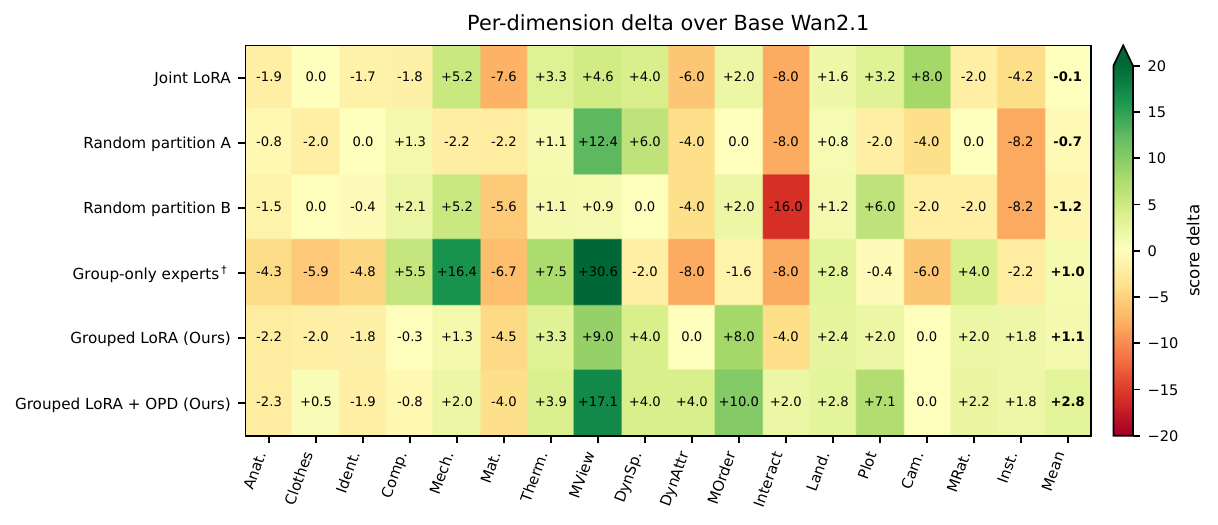}
    \caption{\textbf{Per-dimension score changes over the base model.} Grouped LoRA (Ours) improves multi-view consistency and motion order, while material, human interaction, and camera motion reveal remaining dimension-specific tradeoffs. $^\dagger$Group-only experts are a non-deployable diagnostic: each dimension is scored with the expert of its own group. Colors use a fixed range of $-20$ to $+20$; the upper colorbar extension indicates values above $+20$. Annotations show score changes rounded to one decimal place; the last column is the mean change across 17 dimensions.}
    \label{fig:per_dim_delta}
\end{figure}

Figure~\ref{fig:per_dim_delta} summarizes these dimension-level tradeoffs. Appendix Table~\ref{tab:full_scores} reports the full dimension-level table, including the appendix-only rank-8 merge diagnostic and the OPD adapter.

\begin{figure}[t]
    \centering
    \includegraphics[width=\linewidth]{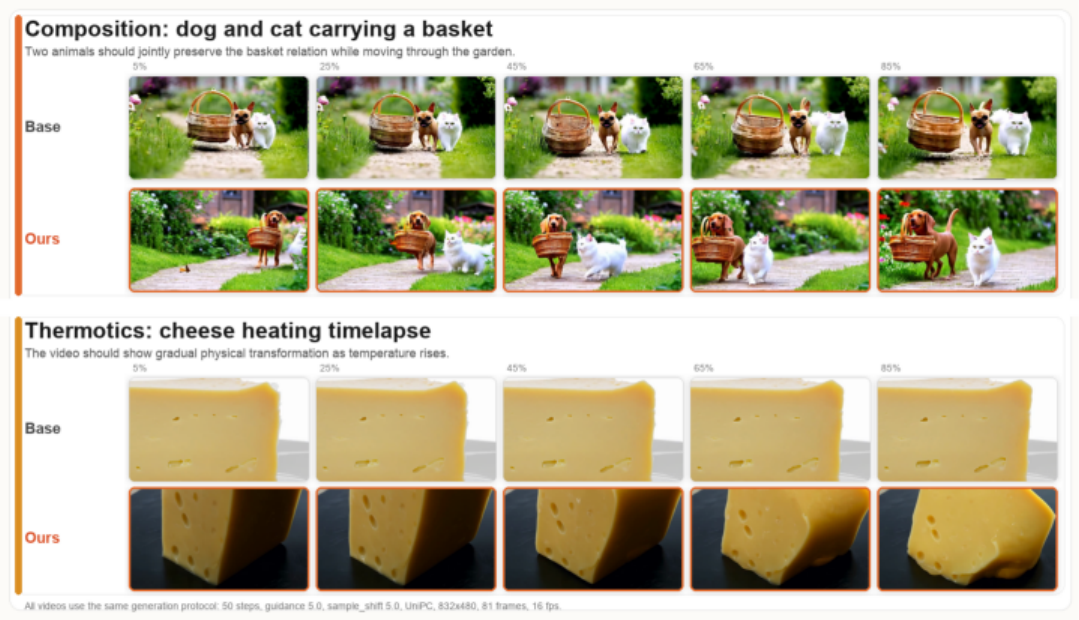}
    \caption{\textbf{Qualitative comparison on representative VBench2.0 dimensions.} We compare Base Wan2.1 and Grouped LoRA (Ours) on Composition and Thermotics prompts; each row shows five temporal frames from one generated video. In the Composition case, Grouped LoRA more consistently preserves the relation among the dog, cat, and basket. In the Thermotics case, it shows clearer signs of cheese softening or melting. These examples are qualitative case studies and complement, rather than replace, the automatic VBench2.0 scores.}
    \label{fig:visual_cases}
\end{figure}

Figure~\ref{fig:visual_cases} illustrates how these dimension-level differences can appear in generated videos. We choose one Composition prompt that requires a dog, a cat, and a basket to remain correctly bound over time, and one Thermotics prompt that requires a visible physical state change under heating. These examples target two different failure modes: relational binding among multiple entities and temporal transformation of material state.

In the Composition example, the base model produces the requested objects but the basket relation is mostly static. Grouped LoRA (Ours) keeps the dog, cat, and basket relation more coherent across the shown frames. In the Thermotics example, the base output remains close to a static cheese block, whereas Grouped LoRA (Ours) shows clearer gradual deformation, matching the intended heat-driven transformation. The cases therefore provide visual evidence for the same pattern seen in the automatic metrics: grouping is most helpful when the target dimension requires preserving a specific relation or transformation rather than merely increasing generic visual quality. They are single cases, however: over all Composition prompts, the score is essentially unchanged (47.80 versus 48.10).

\subsection{Second backbone and independent evaluator}

\begin{table}[t]
    \centering
    \caption{\textbf{Second backbone and independent evaluator.} Left: CogVideoX-2B, VBench2.0 macro score over 17 dimensions ($\times100$). Right: VideoScore-v1.1 five-aspect mean on the same Wan2.1 videos used in Table~\ref{tab:main_results}. Brackets: paired 95\% bootstrap CI.}
    \label{tab:generalization}
    \small
    \begin{tabular}{lr@{\hspace{3em}}lr}
        \toprule
        \multicolumn{2}{c}{CogVideoX-2B (VBench2.0)} & \multicolumn{2}{c}{Wan2.1 (VideoScore-v1.1)} \\
        \cmidrule(r){1-2}\cmidrule(l){3-4}
        Base & 39.87 & Base & 2.914 \\
        Joint LoRA & 27.95 & Grouped LoRA (Ours) & 3.080 \\
        Grouped LoRA (Ours) & 40.99 & & \\
        \midrule
        Ours $-$ Joint & $+13.04$ [9.08, 16.92] & Ours $-$ Base & $+0.166$ [0.140, 0.192] \\
        \bottomrule
    \end{tabular}
\end{table}

On CogVideoX-2B, which changes both the architecture and the training objective, joint training degrades the base model from 39.87 to 27.95, whereas the grouped adapter reaches 40.99 (Table~\ref{tab:generalization}). Most of this margin reflects avoided negative transfer rather than a large gain over the base model (+1.12). The joint degradation is already present at 25\% of training (36.78), so checkpoint selection does not rescue it, and it is dimension-selective: human anatomy, human identity, and thermotics improve while human interaction, motion order, and motion rationality collapse (Appendix~\ref{app:cogvideox}), consistent with capability averaging rather than a globally failed run.

Because the training rewards come from Qwen3-VL, we rescored the Wan2.1 videos with VideoScore-v1.1, a learned metric trained on human ratings of generated videos and not built on Qwen. Grouped LoRA improves all five VideoScore aspects over the base model (Table~\ref{tab:generalization}; Appendix~\ref{app:videoscore}).

\subsection{Dimension-level diagnostics}

To assess whether merging contributes to the remaining dimension-level tradeoffs, we compare each group expert with the merged adapter on its assigned dimensions.

In the matched retention evaluation (Appendix~\ref{app:retention}; group-only row of Figure~\ref{fig:per_dim_delta}), group-only LoRAs reach 48.06 on their assigned dimensions, compared with 48.18 for the merged grouped adapter and 47.07 for the base model. The small average merge gap suggests no global capability collapse, while the larger gaps on multi-view consistency and mechanics indicate that weight-space merging is a plausible source of remaining group-specific tradeoffs. On all six dimensions where the expert beats the base model, the merged adapter is below the expert; on the other eleven, it is above the expert, indicating transfer across groups. Selecting post hoc the better of the two for each dimension gives a diagnostic upper bound of 51.07. Neither rank truncation (error $1.5\times10^{-6}$) nor sign cancellation between expert deltas (no negative pair in 240 modules) explains this loss (Appendix~\ref{app:retention}).

Distilling the experts back into the merged adapter (Eq.~\ref{eq:opd}) raises the mean from 48.18 to 49.91 (+1.73; paired 95\% bootstrap CI $[+0.42, +3.05]$), closing about 60\% of the gap to this upper bound, and improves 12 of 17 dimensions over the base model (Table~\ref{tab:main_results}; last row of Figure~\ref{fig:per_dim_delta}). The largest gains are on multi-view consistency (+8.07), human interaction (+6.00), and complex plot (+5.07). Human interaction reaches 68.00 although its expert scores 58.00, so OPD does not simply copy the teacher. Recovery is uneven: only 0.71 of the 15.14-point mechanics gap is recovered, composition decreases slightly ($-0.50$), and camera motion is unchanged. The interval reflects evaluation sampling for one OPD training run; variation across training runs was not measured.

\section{Related work}

Text-to-video generation builds on diffusion and flow-based modeling~\citep{ho2020denoising,rombach2022latent,peebles2023scalable,blattmann2023align}, with evaluation moving toward multi-axis diagnosis through EvalCrafter, T2V-CompBench, VideoPhy, VBench, and VBench2.0~\citep{liu2023evalcrafter,sun2024t2vcompbench,bansal2024videophy,huang2023vbench,zheng2025vbench2}. We use VBench2.0 dimensions as both metrics and data categories. Diffusion alignment and efficient video generation are reviewed by \citet{liu2026alignment} and \citet{shao2026efficientvideo}, respectively. Zigzag Diffusion Sampling~\citep{bai2025zigzag} and Weak-to-Strong Diffusion with Reflection~\citep{bai2026weak} improve generation through reflective sampling without retraining.

Our training uses LoRA~\citep{hu2021lora} and Qwen3-VL-8B-Instruct QA rewards~\citep{bai2025qwen3vl}, but differs from RLHF/DPO-style language alignment~\citep{bai2022training,rafailov2023direct} and diffusion reward optimization methods~\citep{black2023training,wallace2024diffusion,prabhudesai2023aligning}: the goal is not a new objective, but organizing reward-weighted video data. CRAFT~\citep{sun2026craft} studies composite-reward filtering followed by supervised fine-tuning; our focus is instead on which reward-weighted categories should share updates.

The grouping signal is related to gradient conflict and multi-task optimization methods~\citep{sener2018multi,chen2018gradnorm,yu2020pcgrad,wang2021gradient,liu2021cagrad}, while the merged adapter connects to weight-space merging~\citep{wortsman2022modelsoups,matena2022merging,ilharco2023editing,yadav2023ties}. \method{} intervenes between these lines: it probes residual category gradients before training specialists, then merges the resulting LoRA experts into one deployable adapter. CASA~\citep{wang2026casa} addresses spectral interference when transferring LoRAs to distilled video backbones, whereas we group and merge experts trained on the same backbone.

Our consolidation follows on-policy distillation~\citep{agarwal2024onpolicy} and the multi-teacher paradigm of Flow-OPD~\citep{fang2026flowopd}. Our distinction lies in constructing teachers by residual-gradient grouping of reward-weighted video categories, rather than by single-reward specialization. MaineCoon~\citep{bai2026mainecoon} also consolidates domain-specialized LoRA experts, using preference optimization and reinforced on-policy distillation for streaming audio-visual generation rather than residual-gradient grouping. Our capability-consolidation goal differs from the few-step generation objective of Adaptive Matching Distillation~\citep{bai2026adaptive}.

\section{Limitations}

This study covers two open text-to-video backbones of at most 2B parameters and a VBench2.0-derived category taxonomy; on CogVideoX-2B the gain over the base model is small, so the most natural next step is to test whether the same grouping signal appears across larger model families, reward sources, and evaluation suites. Rewards come from Qwen3-VL and VBench2.0 also uses vision-language judgments; VideoScore agreement reduces but does not remove this coupling, and no human study was run. On Wan2.1, the 95\% interval of the merged adapter's gain over the base model includes zero. OPD adds about 205 GPU-hours and was not applied to the control partitions or to CogVideoX. Our gradient compatibility measure is also intentionally lightweight: it is designed as a practical probe for organizing post-training data rather than as a complete account of all denoising dynamics. Future work could combine this probe with richer merge rules, repeated control partitions, and human preference studies to further separate data-organization effects from evaluator- or merge-specific effects.

\section{Conclusion}

We presented \method{}, a gradient-guided procedure for organizing heterogeneous reward-weighted post-training data. The method computes category gradients, removes the shared global direction, clusters residual compatibility, trains group-specific LoRA experts, and consolidates them by merging and on-policy distillation. The central view is that reward-weighted video SFT induces multiple conditional velocity fields, and post-training should ask not only which samples are high quality, but also which reward-weighted samples should share update directions. VBench2.0 results support this view in a qualified way: Grouped LoRA improves the mean score over both the base model and joint LoRA, while reducing the worst dimension-level drop; it avoids joint training's negative transfer on CogVideoX-2B, and distillation recovers part of the merging loss. Overall, the results position gradient compatibility as a practical organizing signal for reward-weighted video post-training.

\section*{Ethics statement}

The method is a post-training data organization procedure and does not directly introduce a new video generation backbone. Its positive impact is that it may make post-training more data-efficient and more diagnosable by exposing when training categories interfere. Its negative impact follows the underlying model class: improved video generation can also improve misleading or harmful generated media. Any future release of trained adapters or generated videos should therefore follow the safety and licensing constraints of the base model and benchmark assets, and should document whether generated samples are filtered before release.

\section*{Reproducibility statement}

Section~2 and the appendices describe the method, experimental settings, reward construction, and diagnostic protocols. All methods share a prompt manifest and seed schedule for evaluation. We plan to release the implementation, configurations, partitions, prompt manifests, and raw evaluator outputs.

\section*{AI use statement}

Generative AI tools were used to assist with writing and formatting. Their use for data generation, reward construction, distillation, and evaluation is described in the main text and appendices. The authors take responsibility for all claims, results, and artifacts.

\bibliographystyle{iclr2027_conference}
\bibliography{references}

\begin{thebibliography}{41}
\providecommand{\natexlab}[1]{#1}
\providecommand{\url}[1]{\texttt{#1}}
\expandafter\ifx\csname urlstyle\endcsname\relax
  \providecommand{\doi}[1]{doi: #1}\else
  \providecommand{\doi}{doi: \begingroup \urlstyle{rm}\Url}\fi

\bibitem[Agarwal et~al.(2024)Agarwal, Vieillard, Zhou, Stanczyk, Ramos, Geist,
  and Bachem]{agarwal2024onpolicy}
Rishabh Agarwal, Nino Vieillard, Yongchao Zhou, Piotr Stanczyk, Sabela Ramos,
  Matthieu Geist, and Olivier Bachem.
\newblock On-policy distillation of language models: Learning from
  self-generated mistakes.
\newblock In \emph{International Conference on Learning Representations}, 2024.

\bibitem[Bai et~al.(2025{\natexlab{a}})Bai, Shao, Zhou, Qi, Xu, Xiong, and
  Xie]{bai2025zigzag}
Lichen Bai, Shitong Shao, Zikai Zhou, Zipeng Qi, Zhiqiang Xu, Haoyi Xiong, and
  Zeke Xie.
\newblock Zigzag diffusion sampling: Diffusion models can self-improve via
  self-reflection.
\newblock In \emph{International Conference on Learning Representations},
  2025{\natexlab{a}}.
\newblock URL \url{https://openreview.net/forum?id=MKvQH1ekeY}.

\bibitem[Bai et~al.(2026{\natexlab{a}})Bai, Sugiyama, and Xie]{bai2026weak}
Lichen Bai, Masashi Sugiyama, and Zeke Xie.
\newblock Weak-to-strong diffusion with reflection.
\newblock In \emph{International Conference on Learning Representations},
  2026{\natexlab{a}}.
\newblock URL \url{https://openreview.net/forum?id=tg19FVh3p1}.

\bibitem[Bai et~al.(2026{\natexlab{b}})Bai, Zhang, Shao, Tan, Zhong, Xie, Li,
  Huang, Shen, Ji, Wang, Wu, Zhao, Zhu, Luo, Yang, and Xie]{bai2026mainecoon}
Lichen Bai, Tianhao Zhang, Shitong Shao, Dingwei Tan, Qiyu Zhong, Zhengpeng
  Xie, Haopeng Li, Qinghao Huang, Dandan Shen, Tengjiao Ji, Wei Wang, Peicheng
  Wu, Yuxuan Zhao, Xiangyu Zhu, Welly Luo, Shurui Yang, and Zeke Xie.
\newblock {MaineCoon}: Pursuing a real-time audio-visual social world model.
\newblock \emph{arXiv preprint arXiv:2606.17800}, 2026{\natexlab{b}}.
\newblock URL \url{https://arxiv.org/abs/2606.17800}.

\bibitem[Bai et~al.(2026{\natexlab{c}})Bai, Zhou, Shao, Zhong, Yang, Chen,
  Chen, and Xie]{bai2026adaptive}
Lichen Bai, Zikai Zhou, Shitong Shao, Wenliang Zhong, Shuo Yang, Shuo Chen,
  Bojun Chen, and Zeke Xie.
\newblock Optimizing few-step generation with adaptive matching distillation.
\newblock \emph{arXiv preprint arXiv:2602.07345}, 2026{\natexlab{c}}.
\newblock URL \url{https://arxiv.org/abs/2602.07345}.

\bibitem[Bai et~al.(2025{\natexlab{b}})Bai, Cai, Chen, Chen, Chen, Cheng,
  et~al.]{bai2025qwen3vl}
Shuai Bai, Yuxuan Cai, Ruizhe Chen, Keqin Chen, Xionghui Chen, Zesen Cheng,
  et~al.
\newblock {Qwen3-VL} technical report.
\newblock \emph{arXiv preprint arXiv:2511.21631}, 2025{\natexlab{b}}.

\bibitem[Bai et~al.(2022)Bai, Jones, Ndousse, Askell, Chen, DasSarma, Drain,
  Fort, Ganguli, Henighan, et~al.]{bai2022training}
Yuntao Bai, Andy Jones, Kamal Ndousse, Amanda Askell, Anna Chen, Nova DasSarma,
  Dawn Drain, Stanislav Fort, Deep Ganguli, Tom Henighan, et~al.
\newblock Training a helpful and harmless assistant with reinforcement learning
  from human feedback.
\newblock \emph{arXiv preprint arXiv:2204.05862}, 2022.

\bibitem[Bansal et~al.(2024)Bansal, Lin, Xie, Zong, Yarom, Bitton, Jiang, Sun,
  Chang, and Grover]{bansal2024videophy}
Hritik Bansal, Zongyu Lin, Tianyi Xie, Zeshun Zong, Michal Yarom, Yonatan
  Bitton, Chenfanfu Jiang, Yizhou Sun, Kai-Wei Chang, and Aditya Grover.
\newblock Videophy: Evaluating physical commonsense for video generation.
\newblock \emph{arXiv preprint arXiv:2406.03520}, 2024.

\bibitem[Black et~al.(2023)Black, Janner, Du, Kostrikov, and
  Levine]{black2023training}
Kevin Black, Michael Janner, Yilun Du, Ilya Kostrikov, and Sergey Levine.
\newblock Training diffusion models with reinforcement learning.
\newblock \emph{arXiv preprint arXiv:2305.13301}, 2023.

\bibitem[Blattmann et~al.(2023)Blattmann, Rombach, Ling, Dockhorn, Kim, Fidler,
  and Kreis]{blattmann2023align}
Andreas Blattmann, Robin Rombach, Huan Ling, Tim Dockhorn, Seung~Wook Kim,
  Sanja Fidler, and Karsten Kreis.
\newblock Align your latents: High-resolution video synthesis with latent
  diffusion models.
\newblock In \emph{Proceedings of the IEEE/CVF Conference on Computer Vision
  and Pattern Recognition}, pp.\  22563--22575, 2023.
\newblock \doi{10.1109/CVPR52729.2023.02161}.

\bibitem[Chen et~al.(2018)Chen, Badrinarayanan, Lee, and
  Rabinovich]{chen2018gradnorm}
Zhao Chen, Vijay Badrinarayanan, Chen-Yu Lee, and Andrew Rabinovich.
\newblock Gradnorm: Gradient normalization for adaptive loss balancing in deep
  multitask networks.
\newblock In \emph{Proceedings of the 35th International Conference on Machine
  Learning}, volume~80 of \emph{Proceedings of Machine Learning Research}, pp.\
   794--803. PMLR, 2018.

\bibitem[Fang et~al.(2026)Fang, Huang, Zeng, Zhao, Chen, Feng, Lin, Chen, Chen,
  Cao, and Zhao]{fang2026flowopd}
Zhen Fang, Wenxuan Huang, Yu~Zeng, Yiming Zhao, Shuang Chen, Kaituo Feng,
  Yunlong Lin, Lin Chen, Zehui Chen, Shaosheng Cao, and Feng Zhao.
\newblock Flow-{OPD}: On-policy distillation for flow matching models.
\newblock \emph{arXiv preprint arXiv:2605.08063}, 2026.

\bibitem[He et~al.(2024)He, Jiang, Zhang, Ku, Soni, Siu, Chen, Chandra, Jiang,
  Arulraj, Wang, Do, Ni, Lyu, Narsupalli, Fan, Lyu, Lin, and
  Chen]{he2024videoscore}
Xuan He, Dongfu Jiang, Ge~Zhang, Max Ku, Achint Soni, Sherman Siu, Haonan Chen,
  Abhranil Chandra, Ziyan Jiang, Aaran Arulraj, Kai Wang, Quy~Duc Do, Yuansheng
  Ni, Bohan Lyu, Yaswanth Narsupalli, Rongqi Fan, Zhiheng Lyu, Bill~Yuchen Lin,
  and Wenhu Chen.
\newblock {VideoScore}: Building automatic metrics to simulate fine-grained
  human feedback for video generation.
\newblock In \emph{Proceedings of the 2024 Conference on Empirical Methods in
  Natural Language Processing}, 2024.

\bibitem[Ho et~al.(2020)Ho, Jain, and Abbeel]{ho2020denoising}
Jonathan Ho, Ajay Jain, and Pieter Abbeel.
\newblock Denoising diffusion probabilistic models.
\newblock In \emph{Advances in Neural Information Processing Systems},
  volume~33, 2020.
\newblock URL
  \url{https://proceedings.neurips.cc/paper/2020/hash/4c5bcfec8584af0d967f1ab10179ca4b-Abstract.html}.

\bibitem[Hu et~al.(2022)Hu, Shen, Wallis, Allen-Zhu, Li, Wang, Wang, and
  Chen]{hu2021lora}
Edward~J. Hu, Yelong Shen, Phillip Wallis, Zeyuan Allen-Zhu, Yuanzhi Li, Shean
  Wang, Lu~Wang, and Weizhu Chen.
\newblock Lora: Low-rank adaptation of large language models.
\newblock In \emph{International Conference on Learning Representations}, 2022.
\newblock URL \url{https://openreview.net/forum?id=nZeVKeeFYf9}.

\bibitem[Huang et~al.(2024)Huang, He, Yu, Zhang, Si, Jiang, Zhang, Wu, Jin,
  Chanpaisit, Wang, Chen, Wang, Lin, Qiao, and Liu]{huang2023vbench}
Ziqi Huang, Yinan He, Jiashuo Yu, Fan Zhang, Chenyang Si, Yuming Jiang, Yuanhan
  Zhang, Tianxing Wu, Qingyang Jin, Nattapol Chanpaisit, Yaohui Wang, Xinyuan
  Chen, Limin Wang, Dahua Lin, Yu~Qiao, and Ziwei Liu.
\newblock Vbench: Comprehensive benchmark suite for video generative models.
\newblock In \emph{Proceedings of the IEEE/CVF Conference on Computer Vision
  and Pattern Recognition}, pp.\  21807--21818, 2024.
\newblock \doi{10.1109/CVPR52733.2024.02060}.

\bibitem[{Hugging Face}(2024)]{diffusers2024}
{Hugging Face}.
\newblock Diffusers: State-of-the-art diffusion models.
\newblock \url{https://github.com/huggingface/diffusers}, 2024.
\newblock Software library.

\bibitem[Ilharco et~al.(2023)Ilharco, Ribeiro, Wortsman, Gururangan, Schmidt,
  Hajishirzi, and Farhadi]{ilharco2023editing}
Gabriel Ilharco, Marco~Tulio Ribeiro, Mitchell Wortsman, Suchin Gururangan,
  Ludwig Schmidt, Hannaneh Hajishirzi, and Ali Farhadi.
\newblock Editing models with task arithmetic.
\newblock In \emph{International Conference on Learning Representations}, 2023.
\newblock URL \url{https://openreview.net/forum?id=6t0Kwf8-jrj}.

\bibitem[Lipman et~al.(2023)Lipman, Chen, Ben-Hamu, Nickel, and
  Le]{lipman2022flow}
Yaron Lipman, Ricky T.~Q. Chen, Heli Ben-Hamu, Maximilian Nickel, and Matt Le.
\newblock Flow matching for generative modeling.
\newblock In \emph{International Conference on Learning Representations}, 2023.
\newblock URL \url{https://openreview.net/forum?id=PqvMRDCJT9t}.

\bibitem[Liu et~al.(2021)Liu, Liu, Jin, Stone, and Liu]{liu2021cagrad}
Bo~Liu, Xingchao Liu, Xiaojie Jin, Peter Stone, and Qiang Liu.
\newblock Conflict-averse gradient descent for multi-task learning.
\newblock In \emph{Advances in Neural Information Processing Systems},
  volume~34, pp.\  18878--18890, 2021.

\bibitem[Liu et~al.(2026)Liu, Shao, Li, Bai, Xu, Xiong, Kwok, Helal, and
  Xie]{liu2026alignment}
Buhua Liu, Shitong Shao, Bao Li, Lichen Bai, Zhiqiang Xu, Haoyi Xiong, James
  Tin~Yau Kwok, Sumi Helal, and Zeke Xie.
\newblock Alignment of diffusion models: Fundamentals, challenges, and future.
\newblock \emph{ACM Computing Surveys}, 58\penalty0 (9):\penalty0 1--37, 2026.
\newblock \doi{10.1145/3796982}.
\newblock Article 244.

\bibitem[Liu et~al.(2023)Liu, Cun, Liu, Wang, Zhang, Chen, Liu, Zeng, Chan, and
  Shan]{liu2023evalcrafter}
Yaofang Liu, Xiaodong Cun, Xuebo Liu, Xintao Wang, Yong Zhang, Haoxin Chen,
  Yang Liu, Tieyong Zeng, Raymond Chan, and Ying Shan.
\newblock Evalcrafter: Benchmarking and evaluating large video generation
  models.
\newblock \emph{arXiv preprint arXiv:2310.11440}, 2023.

\bibitem[Matena \& Raffel(2022)Matena and Raffel]{matena2022merging}
Michael Matena and Colin Raffel.
\newblock Merging models with fisher-weighted averaging.
\newblock In \emph{Advances in Neural Information Processing Systems},
  volume~35, pp.\  17703--17716, 2022.

\bibitem[Peebles \& Xie(2023)Peebles and Xie]{peebles2023scalable}
William Peebles and Saining Xie.
\newblock Scalable diffusion models with transformers.
\newblock In \emph{Proceedings of the IEEE/CVF International Conference on
  Computer Vision}, pp.\  4195--4205, 2023.
\newblock \doi{10.1109/ICCV51070.2023.00387}.

\bibitem[Prabhudesai et~al.(2023)Prabhudesai, Goyal, Pathak, and
  Fragkiadaki]{prabhudesai2023aligning}
Mihir Prabhudesai, Anirudh Goyal, Deepak Pathak, and Katerina Fragkiadaki.
\newblock Aligning text-to-image diffusion models with reward backpropagation.
\newblock \emph{arXiv preprint arXiv:2310.03739}, 2023.

\bibitem[Rafailov et~al.(2023)Rafailov, Sharma, Mitchell, Manning, Ermon, and
  Finn]{rafailov2023direct}
Rafael Rafailov, Archit Sharma, Eric Mitchell, Christopher~D. Manning, Stefano
  Ermon, and Chelsea Finn.
\newblock Direct preference optimization: Your language model is secretly a
  reward model.
\newblock In \emph{Advances in Neural Information Processing Systems},
  volume~36, pp.\  53728--53741, 2023.

\bibitem[Rombach et~al.(2022)Rombach, Blattmann, Lorenz, Esser, and
  Ommer]{rombach2022latent}
Robin Rombach, Andreas Blattmann, Dominik Lorenz, Patrick Esser, and Bj{\"o}rn
  Ommer.
\newblock High-resolution image synthesis with latent diffusion models.
\newblock In \emph{Proceedings of the IEEE/CVF Conference on Computer Vision
  and Pattern Recognition}, pp.\  10674--10685, 2022.
\newblock \doi{10.1109/CVPR52688.2022.01042}.

\bibitem[Sener \& Koltun(2018)Sener and Koltun]{sener2018multi}
Ozan Sener and Vladlen Koltun.
\newblock Multi-task learning as multi-objective optimization.
\newblock In \emph{Advances in Neural Information Processing Systems},
  volume~31, pp.\  525--536, 2018.

\bibitem[Shao et~al.(2026)Shao, Bai, Wan, Kwok, and
  Xie]{shao2026efficientvideo}
Shitong Shao, Lichen Bai, Pengfei Wan, James Kwok, and Zeke Xie.
\newblock Efficient video diffusion models: Advancements and challenges.
\newblock \emph{arXiv preprint arXiv:2604.15911}, 2026.
\newblock URL \url{https://arxiv.org/abs/2604.15911}.

\bibitem[Sun et~al.(2024)Sun, Huang, Liu, Wu, Xu, Li, and
  Liu]{sun2024t2vcompbench}
Kaiyue Sun, Kaiyi Huang, Xian Liu, Yue Wu, Zihan Xu, Zhenguo Li, and Xihui Liu.
\newblock T2v-compbench: A comprehensive benchmark for compositional
  text-to-video generation.
\newblock \emph{arXiv preprint arXiv:2407.14505}, 2024.

\bibitem[Sun et~al.(2026)Sun, Xie, Bai, Shao, Yang, and Xie]{sun2026craft}
Zening Sun, Zhengpeng Xie, Lichen Bai, Shitong Shao, Shuo Yang, and Zeke Xie.
\newblock {CRAFT}: Aligning diffusion models with fine-tuning is easier than
  you think.
\newblock In \emph{Proceedings of the IEEE/CVF Conference on Computer Vision
  and Pattern Recognition}, 2026.
\newblock URL \url{https://arxiv.org/abs/2603.18991}.

\bibitem[Wallace et~al.(2024)Wallace, Dang, Rafailov, Zhou, Lou, Purushwalkam,
  Ermon, Xiong, Joty, and Naik]{wallace2024diffusion}
Bram Wallace, Meihua Dang, Rafael Rafailov, Linqi Zhou, Aaron Lou, Senthil
  Purushwalkam, Stefano Ermon, Caiming Xiong, Shafiq Joty, and Nikhil Naik.
\newblock Diffusion model alignment using direct preference optimization.
\newblock In \emph{Proceedings of the IEEE/CVF Conference on Computer Vision
  and Pattern Recognition}, pp.\  8228--8238, 2024.
\newblock \doi{10.1109/CVPR52733.2024.00786}.

\bibitem[{Wan-AI}(2025)]{wan2025huggingface}
{Wan-AI}.
\newblock Wan2.1-t2v-1.3b-diffusers.
\newblock \url{https://huggingface.co/Wan-AI/Wan2.1-T2V-1.3B-Diffusers}, 2025.
\newblock Hugging Face model card.

\bibitem[Wang et~al.(2026)Wang, Zhong, Bai, Zhou, Shao, Cheng, Chen, Yang, and
  Xie]{wang2026casa}
Yuchen Wang, Wenliang Zhong, Lichen Bai, Zikai Zhou, Shitong Shao, Bojun Cheng,
  Shuo Chen, Shuo Yang, and Zeke Xie.
\newblock Exploring data-free {LoRA} transferability for video diffusion
  models.
\newblock In \emph{International Conference on Machine Learning}, 2026.
\newblock URL \url{https://arxiv.org/abs/2605.01929}.

\bibitem[Wang et~al.(2021)Wang, Tsvetkov, Firat, and Cao]{wang2021gradient}
Zirui Wang, Yulia Tsvetkov, Orhan Firat, and Yuan Cao.
\newblock Gradient vaccine: Investigating and improving multi-task optimization
  in massively multilingual models.
\newblock In \emph{International Conference on Learning Representations}, 2021.
\newblock URL \url{https://openreview.net/forum?id=F1vEjWK-lH_}.

\bibitem[{WanTeam} et~al.(2025){WanTeam}, Wang, Ai, Wen, Mao, Xie, Chen, Yu,
  Zhao, Yang, et~al.]{wan2025}
{WanTeam}, Ang Wang, Baole Ai, Bin Wen, Chaojie Mao, Chen-Wei Xie, Di~Chen,
  Feiwu Yu, Haiming Zhao, Jianxiao Yang, et~al.
\newblock Wan: Open and advanced large-scale video generative models.
\newblock \emph{arXiv preprint arXiv:2503.20314}, 2025.
\newblock \doi{10.48550/arXiv.2503.20314}.

\bibitem[Wortsman et~al.(2022)Wortsman, Ilharco, Gadre, Roelofs, Gontijo-Lopes,
  Morcos, Namkoong, Farhadi, Carmon, Kornblith, and
  Schmidt]{wortsman2022modelsoups}
Mitchell Wortsman, Gabriel Ilharco, Samir~Yitzhak Gadre, Rebecca Roelofs,
  Raphael Gontijo-Lopes, Ari~S. Morcos, Hongseok Namkoong, Ali Farhadi, Yair
  Carmon, Simon Kornblith, and Ludwig Schmidt.
\newblock Model soups: Averaging weights of multiple fine-tuned models improves
  accuracy without increasing inference time.
\newblock In \emph{Proceedings of the 39th International Conference on Machine
  Learning}, volume 162 of \emph{Proceedings of Machine Learning Research},
  pp.\  23965--23998. PMLR, 2022.

\bibitem[Yadav et~al.(2023)Yadav, Tam, Choshen, Raffel, and
  Bansal]{yadav2023ties}
Prateek Yadav, Derek Tam, Leshem Choshen, Colin~A. Raffel, and Mohit Bansal.
\newblock {TIES}-merging: Resolving interference when merging models.
\newblock In \emph{Advances in Neural Information Processing Systems},
  volume~36, 2023.

\bibitem[Yang et~al.(2025)Yang, Teng, Zheng, Ding, Huang, Xu, Yang, Hong,
  Zhang, Feng, Yin, Zhang, Wang, Cheng, Xu, Gu, Dong, and
  Tang]{yang2025cogvideox}
Zhuoyi Yang, Jiayan Teng, Wendi Zheng, Ming Ding, Shiyu Huang, Jiazheng Xu,
  Yuanming Yang, Wenyi Hong, Xiaohan Zhang, Guanyu Feng, Da~Yin, Yuxuan Zhang,
  Weihan Wang, Yean Cheng, Bin Xu, Xiaotao Gu, Yuxiao Dong, and Jie Tang.
\newblock {CogVideoX}: Text-to-video diffusion models with an expert
  transformer.
\newblock In \emph{International Conference on Learning Representations}, 2025.

\bibitem[Yu et~al.(2020)Yu, Kumar, Gupta, Levine, Hausman, and
  Finn]{yu2020pcgrad}
Tianhe Yu, Saurabh Kumar, Abhishek Gupta, Sergey Levine, Karol Hausman, and
  Chelsea Finn.
\newblock Gradient surgery for multi-task learning.
\newblock In \emph{Advances in Neural Information Processing Systems},
  volume~33, pp.\  5824--5836, 2020.

\bibitem[Zheng et~al.(2025)Zheng, Huang, Liu, Zou, He, Zhang,
  et~al.]{zheng2025vbench2}
Dian Zheng, Ziqi Huang, Hongbo Liu, Kai Zou, Yinan He, Fan Zhang, et~al.
\newblock Vbench-2.0: Advancing video generation benchmark suite for intrinsic
  faithfulness.
\newblock \emph{arXiv preprint arXiv:2503.21755}, 2025.

\end{thebibliography}

\clearpage
\appendix
\let\oldsection\section
\renewcommand{\section}{\FloatBarrier\oldsection}

\section{Training, distillation, and evaluation parameters}
\label{app:training_eval_params}

The main LoRA comparisons use a quick VBench2.0 protocol with 25 prompts per dimension and two generated videos per prompt. This matched protocol is used to compare adapters under identical generation and evaluator settings while keeping evaluation cost practical; the same quick protocol is used consistently for the base, joint, partition-control, grouped, and OPD comparisons.

\begin{table}[h]
\centering
\caption{Generation parameters for the quick VBench2.0 comparisons.}
\begin{tabular}{ll}
\toprule
Parameter & Value \\
\midrule
Base model & Wan2.1-T2V-1.3B-Diffusers \\
Scheduler & UniPC \\
Steps & 50 \\
Guidance scale & 5.0 \\
Sample/flow shift & 5.0 \\
Resolution & $832\times480$ \\
Frames / FPS & 81 / 16 \\
Seed & 42, with per-video seed $42+\mathrm{index}$ \\
Precision & float16 \\
\bottomrule
\end{tabular}
\end{table}

\begin{table}[h]
\centering
\caption{LoRA post-training parameters.}
\begin{tabular}{ll}
\toprule
Item & Setting \\
\midrule
Trainable modules & q, k, v, output projection LoRA \\
Objective & reward-weighted flow matching MSE \\
Reward normalization & sample weight / global mean weight \\
Text length & 512 tokens \\
Probe samples & 100 per dimension for the full gradient-probe run \\
Probe stages & 0/25/50/75/100\% \\
Group count & 5 \\
Group experts & rank 8, one epoch over the group's samples \\
Merged adapter & group-size-weighted dense merge, rank-40 SVD \\
\bottomrule
\end{tabular}
\end{table}

\begin{table}[h]
\centering
\caption{On-policy distillation (OPD) settings on Wan2.1.}
\label{tab:opd_params}
\begin{tabular}{ll}
\toprule
Item & Setting \\
\midrule
Student initialization & merged rank-40 adapter $\phi_0$ (student rank 40) \\
Anchor & frozen merged adapter $\phi_0$ \\
Teachers & five frozen group experts, routed by dimension \\
Prompts & training prompts of the reward pool (videos, weights unused) \\
Loss weights & $\lambda_{\mathrm{OPD}}=1.0$, $\lambda=0.1$ (Eq.~\ref{eq:opd}) \\
Student rollout & 8 sampler steps from fresh noise, no gradient \\
Distilled states & 2 interior rollout states per prompt \\
Guidance during OPD & 1.0 (no classifier-free guidance) \\
Resolution / frames & $832\times480$ / 81 \\
Optimizer & AdamW, learning rate $1\times10^{-6}$, one pass over the prompts \\
Batch & 1 prompt per GPU on 16 GPUs \\
Precision & float16, flow shift 5.0, text length 512 \\
\bottomrule
\end{tabular}
\end{table}

\begin{table}[h]
\centering
\caption{Compute. GPU-hours are measured on the listed hardware.}
\label{tab:compute}
\begin{tabular}{llrl}
\toprule
Stage & Hardware & GPU-hours & Note \\
\midrule
Joint training and gradient probes & NVIDIA A800 & 13.4 & \\
Five group experts & NVIDIA A800 & 33.5 & \\
OPD consolidation & NVIDIA RTX 4090 & $\approx$205 & 16 GPUs, 12 h 48 min \\
\bottomrule
\end{tabular}
\end{table}

\section{Qwen-based reward construction}
\label{app:reward_pipeline}

The reward pipeline converts each generated training video into a scalar sample weight. It uses Qwen3-VL-8B-Instruct in two stages. First, for each prompt and VBench2.0 dimension, Qwen generates a small set of dimension-specific yes/no QA items from a prompt template. Second, Qwen watches the generated video and answers each QA item with either ``Yes'' or ``No''. A sample receives credit for a QA item when the video answer matches the expected answer generated from the text prompt.

For Motion Order Understanding, the QA-generation template instructs Qwen to identify the main subject's core action events and produce two to four yes/no questions about their temporal order. For example, if the prompt describes a subject clapping, bending down, and then picking up a bag, the generated QA items should ask whether the earlier action happens before the later action, or whether the later action starts after the earlier action finishes. The template explicitly excludes camera movement, shot changes, clothing, background, and other non-action properties so that the reward targets event order rather than general visual quality.

Let a video sample produce $N$ valid QA items. For item $j$, let $a_j\in\{0,1\}$ indicate whether Qwen's video answer matches the expected answer. The video-level reward is defined as
\begin{equation}
    r = \frac{1}{N}\sum_{j=1}^{N} a_j.
    \label{eq:qwen_reward}
\end{equation}
The reward is then converted into the training weight used in Eq.~\ref{eq:reward_loss} as
\begin{equation}
    w = \mathrm{clip}\left(1+\alpha(r-\bar{r}), w_{\min}, w_{\max}\right),
    \label{eq:qwen_weight}
\end{equation}
where $\bar{r}$ is the mean reward over retained samples in the same run. The implementation uses $\alpha=0.5$, $w_{\min}=0.5$, and $w_{\max}=2.0$. Samples with no valid QA items are skipped; samples with fewer than the recommended number of QA items are retained with a warning if at least one valid QA item remains.

\paragraph{Reward audit.} Table~\ref{tab:reward_audit} summarizes the Wan2.1 reward pool. Stored weights match Eq.~\ref{eq:qwen_weight}, and all 17 dimensions are covered. Camera motion has the largest loss from low-QA filtering (512 of 560 retained).

\begin{table}[h]
\centering
\caption{Audit of the Wan2.1 reward pipeline.}
\label{tab:reward_audit}
\begin{tabular}{lr}
\toprule
Item & Value \\
\midrule
Produced reward records & 9,512 \\
Records retained for training & 9,448 (99.33\%) \\
Reward-pipeline failures & 0 \\
Duplicate video paths & 0 \\
QA-structure anomalies & 0 \\
Samples at a weight-clipping bound & 0 \\
\bottomrule
\end{tabular}
\end{table}

\section{Partition controls}
\label{app:control_grouping}

The random partition controls use the same number of groups as \method{} but are not selected by the final residual-gradient compatibility criterion. They use the same grouped-LoRA training and merge pipeline as \method{} and are intended to check whether the final result follows from the specific residual-gradient partition rather than from the existence of five smaller adapters alone.

Random partition A uses groups \{complex landscape, complex plot, composition, human identity, mechanics\}, \{dynamic attribute, material, thermotics\}, \{camera motion, human anatomy, multi-view consistency\}, \{motion order, motion rationality, human clothes\}, and \{dynamic spatial relationship, human interaction, instance preservation\}.

Random partition B uses groups \{dynamic spatial relationship, human anatomy, human clothes, human identity, human interaction, instance preservation, motion order, motion rationality\}, \{dynamic attribute, material, mechanics, thermotics\}, \{camera motion, multi-view consistency\}, \{complex landscape, composition\}, and \{complex plot\}.

The semantic partition uses the official VBench2.0 categories without the diversity dimension: Human Fidelity \{human anatomy, human identity, human clothes\}; Creativity \{composition\}; Controllability \{dynamic spatial relationship, dynamic attribute, motion order, human interaction, complex landscape, complex plot, camera motion\}; Physics \{mechanics, thermotics, material, multi-view consistency\}; and Commonsense \{motion rationality, instance preservation\}.

The raw-gradient partition applies average-linkage clustering with $M=5$ to the raw category-gradient cosine matrix, without removing the global direction. Directly subtracting the global gradient instead of projecting it yields exactly the \method{} partition.

\section{Dimension-level VBench2.0 table}

\begin{table}[h]
\centering
\caption{Dimension-level scores under the quick VBench2.0 evaluation protocol. Scores are percentages; ties are counted as neither improved nor degraded in aggregate tables. Grouped r8 is included as an appendix-only merge-rank diagnostic; the main comparison uses Grouped r40 (Ours). Group-only scores each dimension with the expert of its own group (non-deployable diagnostic). +OPD is Grouped r40 after on-policy distillation.}
\resizebox{\linewidth}{!}{%
\begin{tabular}{lrrrrrrrr}
\toprule
Dimension & Base & Joint LoRA & Random A & Random B & Group-only & Grouped r8 & Grouped r40 (Ours) & +OPD (Ours) \\
\midrule
Human Anatomy & 88.21 & 86.27 & 87.38 & 86.66 & 83.88 & 85.00 & 85.99 & 85.94 \\
Human Clothes & 97.92 & 97.96 & 95.92 & 97.96 & 92.00 & 83.33 & 95.92 & 98.40 \\
Human Identity & 73.37 & 71.69 & 73.40 & 73.01 & 68.54 & 77.15 & 71.59 & 71.49 \\
Composition & 48.10 & 46.30 & 49.40 & 50.20 & 53.60 & 46.30 & 47.80 & 47.30 \\
Mechanics & 56.52 & 61.70 & 54.35 & 61.70 & 72.92 & 70.00 & 57.78 & 58.49 \\
Material & 36.67 & 29.03 & 34.48 & 31.03 & 30.00 & 45.83 & 32.14 & 32.70 \\
Thermotics & 47.83 & 51.16 & 48.89 & 48.94 & 55.32 & 66.67 & 51.11 & 51.69 \\
Multi-view Consistency & 42.96 & 47.60 & 55.37 & 43.91 & 73.56 & 30.54 & 51.95 & 60.02 \\
Dynamic Spatial Relationship & 34.00 & 38.00 & 40.00 & 34.00 & 32.00 & 30.00 & 38.00 & 38.00 \\
Dynamic Attribute & 20.00 & 14.00 & 16.00 & 16.00 & 12.00 & 16.00 & 20.00 & 24.00 \\
Motion Order Understanding & 24.00 & 26.00 & 24.00 & 26.00 & 22.45 & 28.57 & 32.00 & 34.04 \\
Human Interaction & 66.00 & 58.00 & 58.00 & 50.00 & 58.00 & 60.03 & 62.00 & 68.00 \\
Complex Landscape & 20.00 & 21.60 & 20.80 & 21.20 & 22.80 & 18.80 & 22.40 & 22.80 \\
Complex Plot & 10.40 & 13.60 & 8.40 & 16.40 & 10.00 & 9.20 & 12.40 & 17.47 \\
Camera Motion & 12.00 & 20.00 & 8.00 & 10.00 & 6.00 & 16.00 & 12.00 & 12.00 \\
Motion Rationality & 38.00 & 36.00 & 38.00 & 36.00 & 42.00 & 38.00 & 40.00 & 40.20 \\
Instance Preservation & 84.21 & 80.00 & 76.00 & 76.00 & 82.00 & 84.51 & 86.00 & 86.00 \\
\midrule
Mean & 47.07 & 46.99 & 46.38 & 45.82 & 48.06 & 47.41 & 48.18 & 49.91 \\
\bottomrule
\end{tabular}%
}
\label{tab:full_scores}
\end{table}

The rank-8 merge is lower than the rank-40 merge on the average score (47.41 vs. 48.18) and is included only to diagnose merge-rank sensitivity. Its dimension-level pattern is uneven: it improves mechanics, material, and thermotics, but drops multi-view consistency and human clothes consistency sharply. This supports using Grouped r40 (Ours) as the main grouped adapter in Table~\ref{tab:main_results}.

\section{Gradient compatibility diagnostics}
\label{app:gradient_diagnostics}

\begin{figure}[h]
    \centering
    \includegraphics[width=\linewidth]{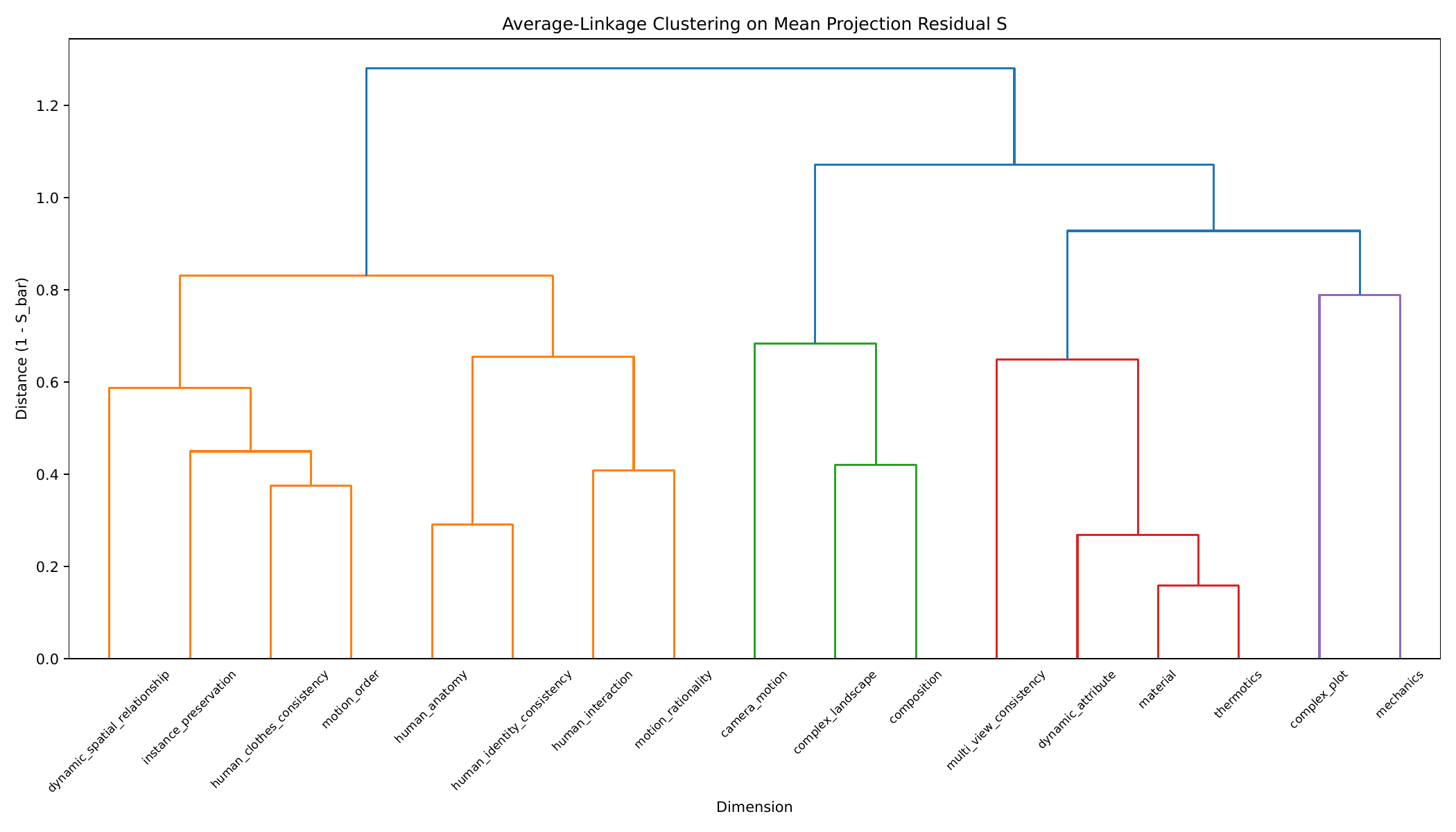}
    \caption{\textbf{Hierarchical clustering over mean residual-gradient compatibility.} The $M=5$ cut separates five groups with positive within-group residual similarity and avoids negative within-group pairs.}
    \label{fig:app_dendrogram}
\end{figure}

Figure~\ref{fig:app_dendrogram} shows the average-linkage clustering tree used to obtain the final grouped LoRA data split. The distance is $1-\bar{S}$, where $\bar{S}$ is the mean projection-residual gradient similarity over five probe stages. This figure complements the heatmap in the main text by showing the hierarchical structure before cutting the tree into five groups.

\begin{figure}[p]
    \centering
    \begin{minipage}{0.48\linewidth}
        \centering
        \includegraphics[width=\linewidth]{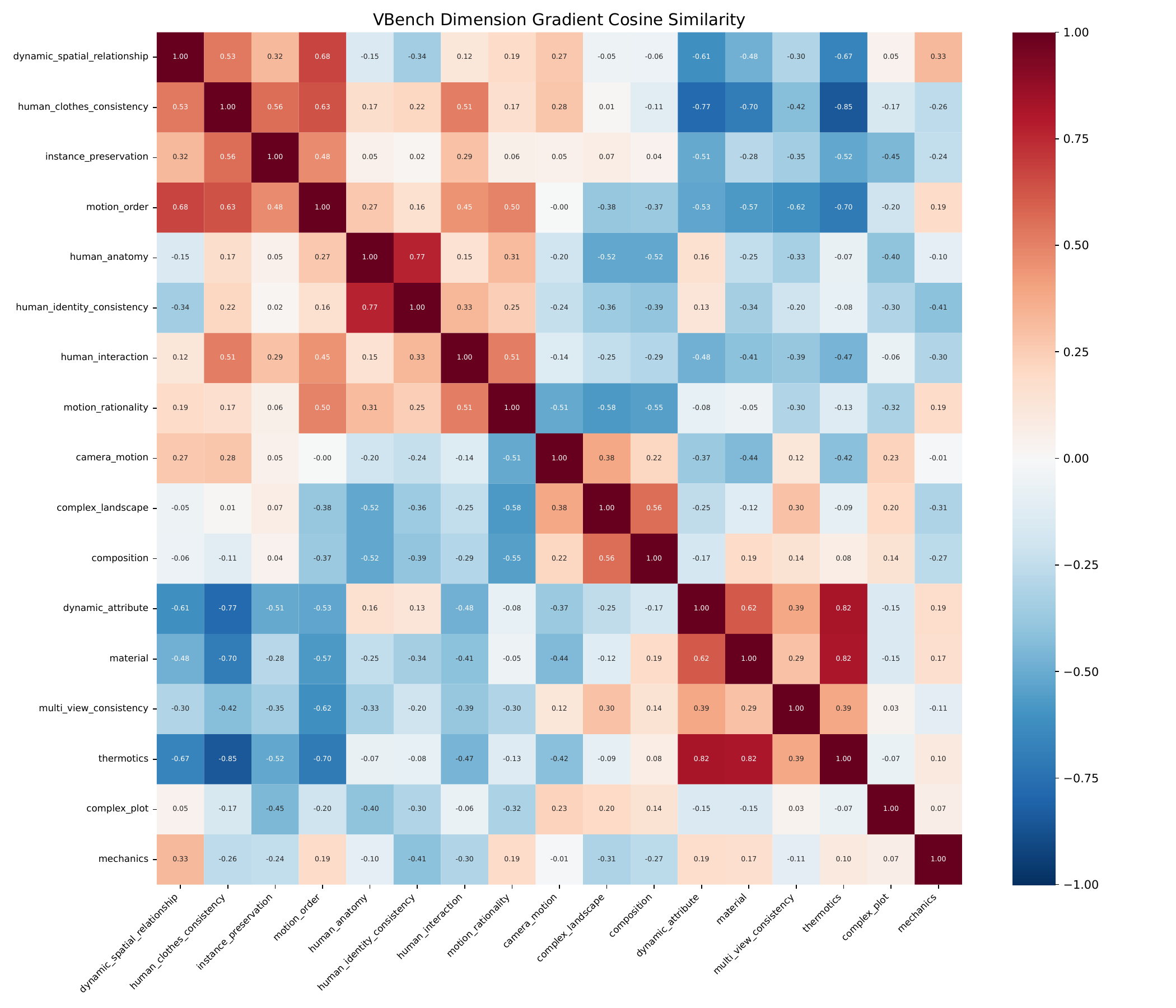}\\
        \small 0\%
    \end{minipage}
    \hfill
    \begin{minipage}{0.48\linewidth}
        \centering
        \includegraphics[width=\linewidth]{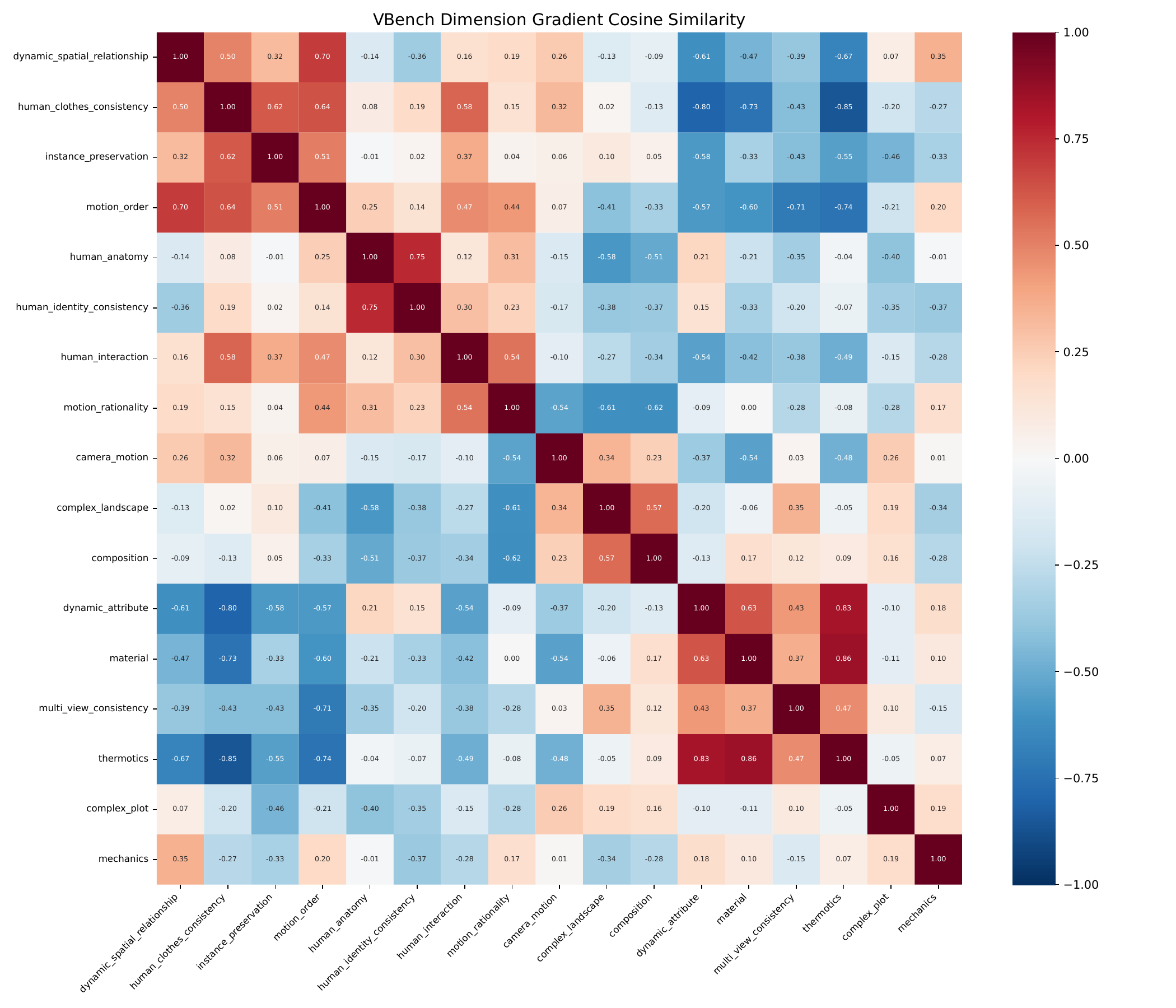}\\
        \small 25\%
    \end{minipage}
    \vspace{0.5em}

    \begin{minipage}{0.48\linewidth}
        \centering
        \includegraphics[width=\linewidth]{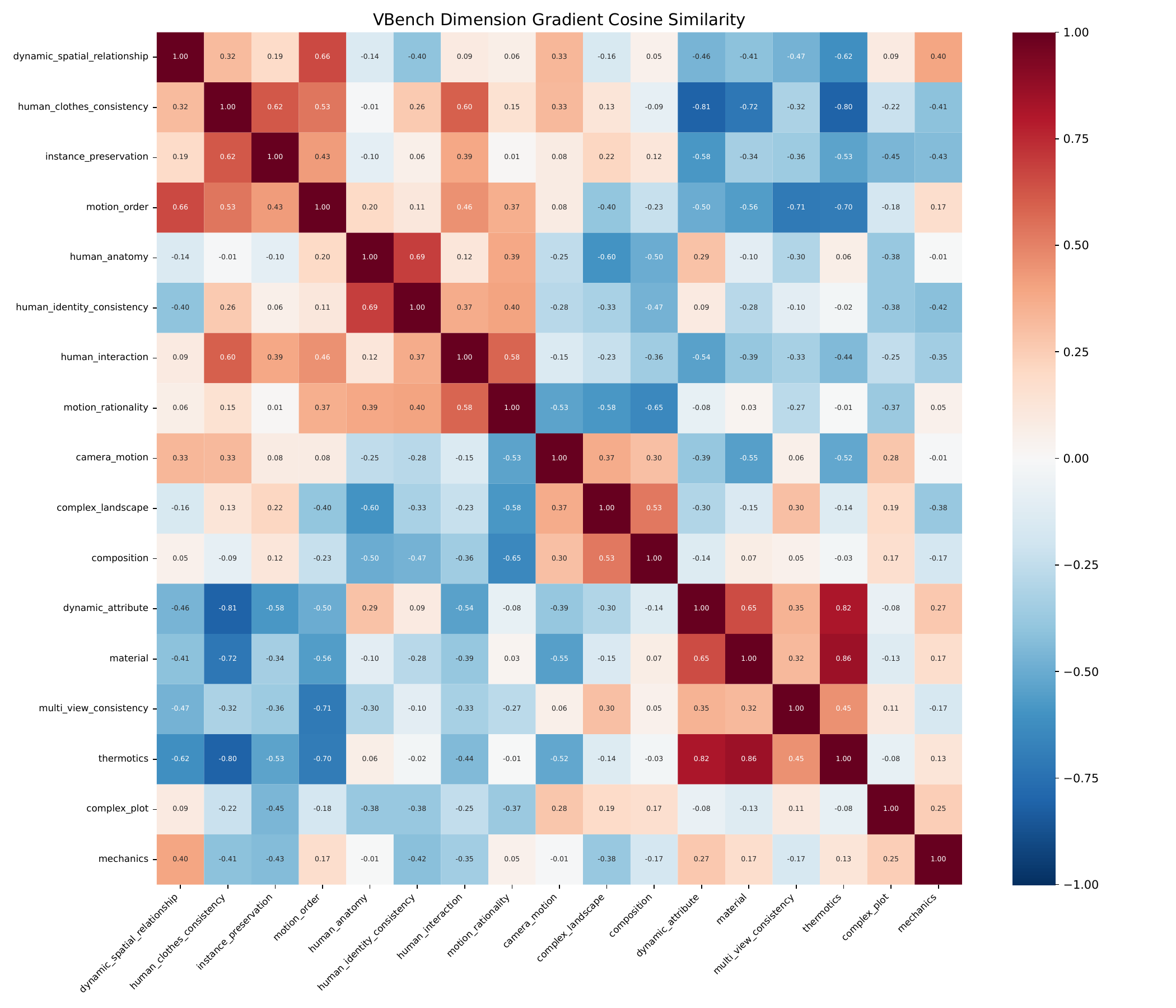}\\
        \small 50\%
    \end{minipage}
    \hfill
    \begin{minipage}{0.48\linewidth}
        \centering
        \includegraphics[width=\linewidth]{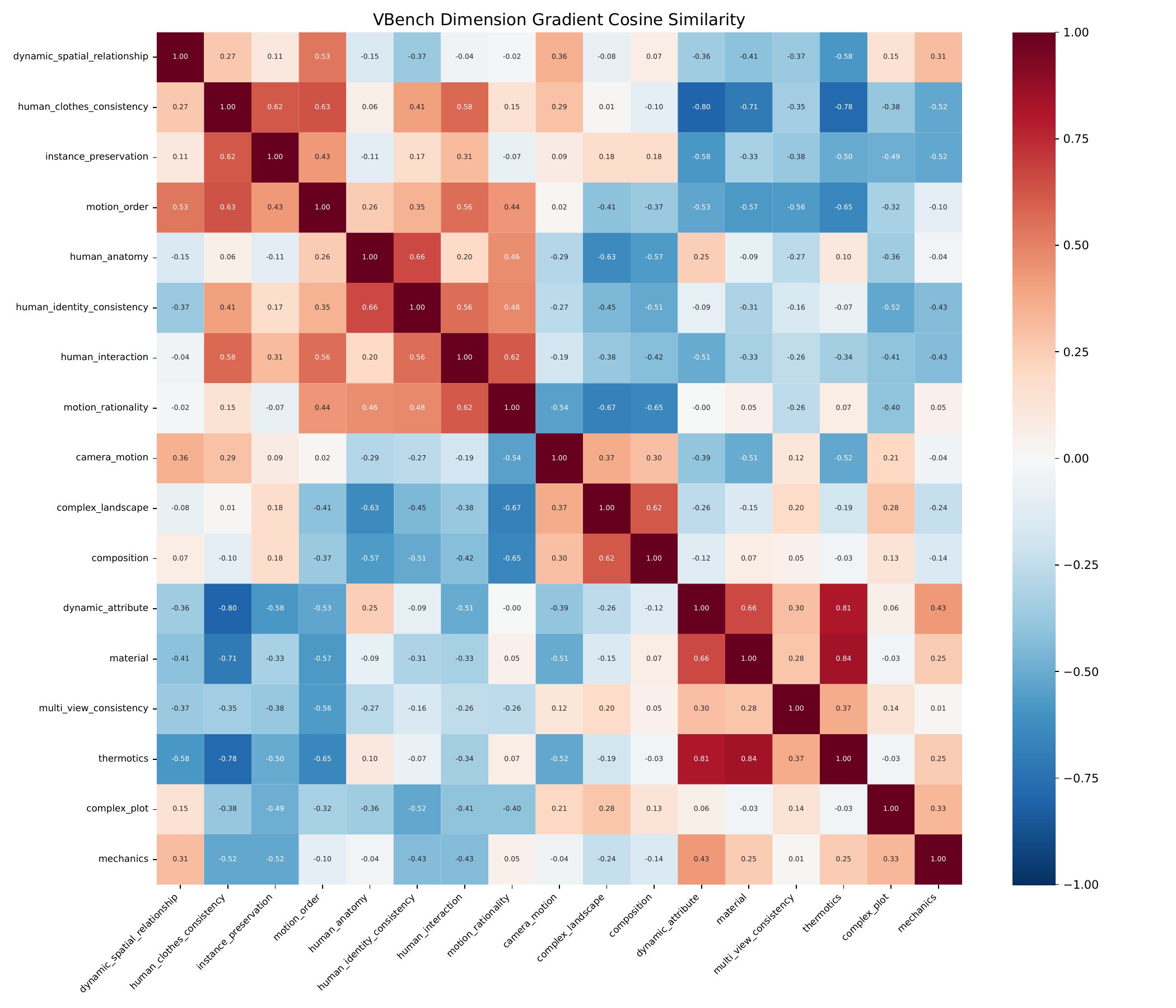}\\
        \small 75\%
    \end{minipage}
    \vspace{0.5em}

    \begin{minipage}{0.48\linewidth}
        \centering
        \includegraphics[width=\linewidth]{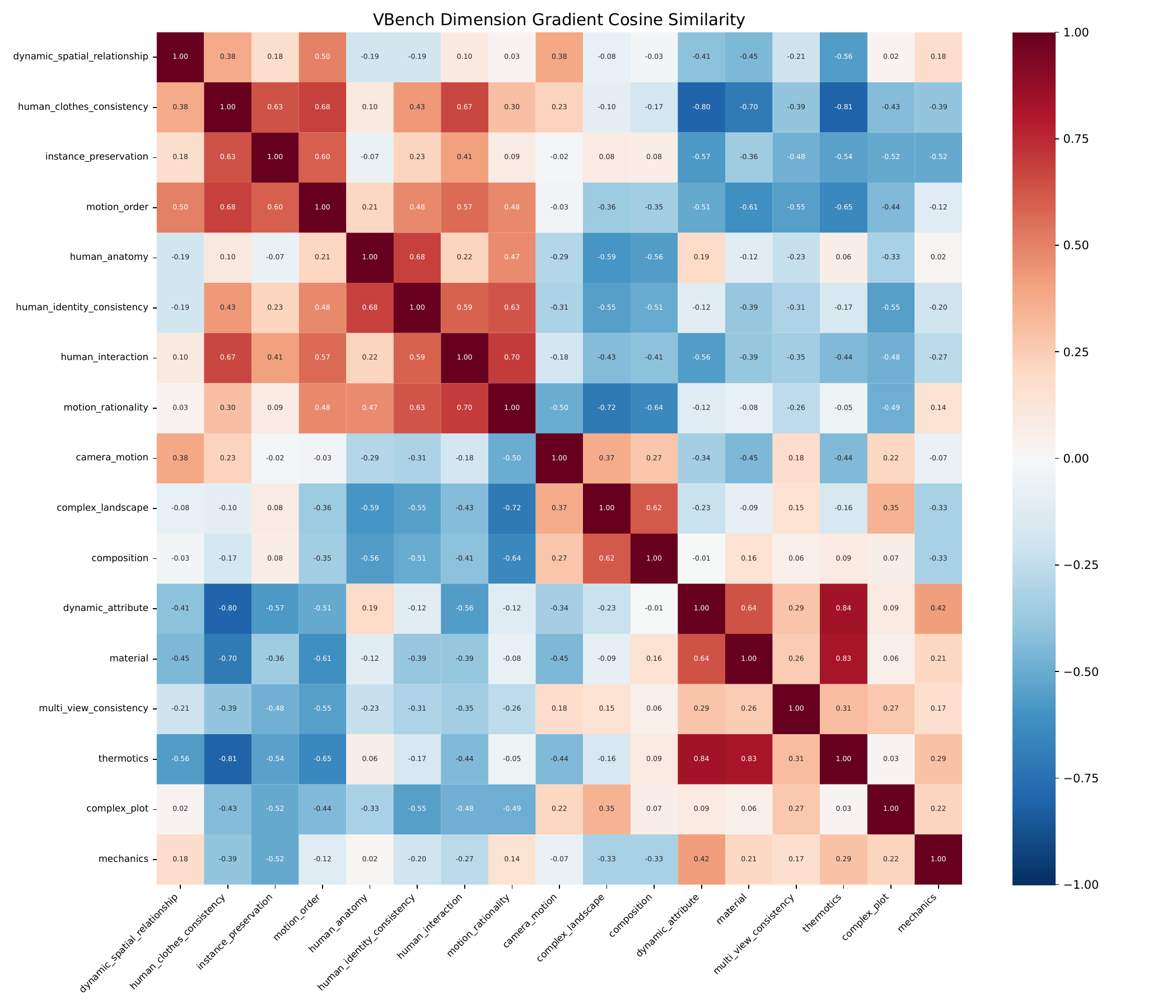}\\
        \small 100\%
    \end{minipage}
    \caption{\textbf{Stage-wise residual-gradient compatibility.} Each heatmap is computed after projecting out the global mixed-gradient direction at the corresponding probe stage. The repeated block structure provides evidence that the grouping is not determined by a single noisy probe.}
\label{fig:app_stage_heatmaps}
\end{figure}

Figure~\ref{fig:app_stage_heatmaps} shows the stage-wise projection-residual similarity matrices at 0\%, 25\%, 50\%, 75\%, and 100\% of the reference joint training trajectory. All five heatmaps use the final $M=5$ grouping order. Their role is diagnostic: they indicate whether the compatibility structure used for grouping is a persistent training signal rather than a one-off probe artifact.

\paragraph{Shared-direction removal.} Table~\ref{tab:signal} compares the raw gradient, direct subtraction of the global gradient, and the projection residual of Eq.~\ref{eq:residual}. Raw gradients make every pair positive. Direct subtraction and projection give nearly identical matrices (Pearson/Spearman 0.998/0.997) and the same $M=5$ partition (adjusted Rand index 1.0), so they define the same downstream model.

\begin{table}[h]
\centering
\caption{Grouping signals on Wan2.1. Separation is the within-group minus between-group mean similarity of the resulting $M=5$ partition, computed on the corresponding matrix.}
\label{tab:signal}
\begin{tabular}{lrrr}
\toprule
Signal & Negative pairs & Separation & VBench2.0 mean \\
\midrule
Raw gradient & 0/136 & 0.046 & 47.35 \\
Direct subtraction & 76/136 & 0.635 & 48.18 (same partition) \\
Projection residual (Eq.~\ref{eq:residual}) & 75/136 & 0.633 & 48.18 \\
\bottomrule
\end{tabular}
\end{table}

\paragraph{Choice of the number of groups.} Table~\ref{tab:k_select} compares $M=4$ and $M=5$ on the mean residual matrix. The choice uses training gradients only, not evaluation scores.

\begin{table}[h]
\centering
\caption{Clustering diagnostics for the number of groups on Wan2.1.}
\label{tab:k_select}
\begin{tabular}{lrrr}
\toprule
Groups & Within-group mean & Separation & Negative within-group pairs \\
\midrule
$M=4$ & 0.345 & 0.559 & 3 \\
$M=5$ & 0.473 & 0.633 & 0 \\
\bottomrule
\end{tabular}
\end{table}

\paragraph{Prevalence across backbones and probe cohorts.} Table~\ref{tab:prevalence} reports residual conflict for the Wan2.1 probes and for three non-overlapping CogVideoX-2B probe cohorts. On Wan2.1, 65 of 136 pairs are negative at all five training stages and 71 at four or more, and the stage-wise matrices have mean Pearson/Spearman correlation 0.966/0.962.

\begin{table}[h]
\centering
\caption{Residual-gradient conflict across backbones and probe cohorts.}
\label{tab:prevalence}
\begin{tabular}{lrrrr}
\toprule
Backbone / probe cohort & Negative pairs & Within-group & Between-group & Separation \\
\midrule
Wan2.1 probes & 75/136 (55.1\%) & $+0.473$ & $-0.160$ & $+0.633$ \\
CogVideoX-2B main probes & 84/136 (61.8\%) & $+0.154$ & $-0.102$ & $+0.257$ \\
CogVideoX-2B cohort A & 81/136 (59.6\%) & $+0.177$ & $-0.121$ & $+0.298$ \\
CogVideoX-2B cohort B & 87/136 (64.0\%) & $+0.191$ & $-0.121$ & $+0.312$ \\
\bottomrule
\end{tabular}
\end{table}

\paragraph{Probe resampling.} Three probe manifests, each independently sampling 100 examples per dimension with no overlap with held-out examples, were compared with the full reference matrix at the final Wan2.1 checkpoint (Table~\ref{tab:resample}). Strong relations are the most reproducible.

\begin{table}[h]
\centering
\caption{Stability of the projection-residual matrix under probe resampling on Wan2.1.}
\label{tab:resample}
\begin{tabular}{lrrrr}
\toprule
Comparison & Pearson & Spearman & Sign agr.\ (all) & Sign agr.\ (strong) \\
\midrule
Resamples vs.\ full reference & 0.847 & 0.834 & 0.816 & 0.853 \\
Across the three resamples & 0.728 & 0.703 & 0.755 & 0.787 \\
\bottomrule
\end{tabular}
\end{table}

\paragraph{Held-out transfer.} Gradients were estimated from 100 probe samples per dimension; cross-dimension loss changes after a small step along each category gradient were measured on a disjoint held-out set. Table~\ref{tab:transfer} reports the rank correlation between residual similarity and held-out transfer. The transfer rankings are nearly identical across step sizes (pairwise Spearman 0.9992--0.9997).

\begin{table}[h]
\centering
\caption{Residual similarity predicts held-out cross-dimension transfer on Wan2.1.}
\label{tab:transfer}
\begin{tabular}{rrr}
\toprule
Relative step size & Spearman $\rho$ & Permutation $p$ \\
\midrule
0.0005 & 0.256 & 0.0028 \\
0.001 & 0.255 & 0.0038 \\
0.002 & 0.252 & 0.0028 \\
\bottomrule
\end{tabular}
\end{table}

\section{Dense delta-SVD LoRA merge}
\label{app:merge}

The grouped rank-40 adapter is produced by merging the five group-specific LoRA experts in dense weight-delta space and then recompressing the result back into LoRA factors. For a linear layer with frozen base weight $W$ and a LoRA expert $m$, let $A_m\in\mathbb{R}^{r_m\times d_{\mathrm{in}}}$ and $B_m\in\mathbb{R}^{d_{\mathrm{out}}\times r_m}$ denote the learned LoRA factors. The dense update represented by the expert is defined as
\begin{equation}
    \Delta W_m = B_m A_m,
    \label{eq:dense_delta}
\end{equation}
where the implementation uses unit LoRA scaling for the merge. The group experts are then averaged in dense delta space as
\begin{equation}
    \Delta W_{\mathrm{merge}} = \sum_{m=1}^{M} \alpha_m \Delta W_m,\qquad \sum_{m=1}^{M}\alpha_m=1.
    \label{eq:dense_merge}
\end{equation}
For the rank-40 grouped adapter, the weights are proportional to the number of VBench2.0 dimensions assigned to each group. With group sizes $[4,4,3,4,2]$, the corresponding weights are $[4/17,4/17,3/17,4/17,2/17]$.

The dense merged update is converted back into a rank-$r$ LoRA module through truncated singular value decomposition. For each merged layer, this is computed as
\begin{equation}
    \Delta W_{\mathrm{merge}} \approx U_r \Sigma_r V_r^\top,
    \label{eq:truncated_svd}
\end{equation}
with requested rank $r=40$. The recompressed LoRA factors are defined as
\begin{equation}
    B_{\mathrm{new}} = U_r \Sigma_r^{1/2},\qquad A_{\mathrm{new}} = \Sigma_r^{1/2} V_r^\top.
    \label{eq:lora_recompose}
\end{equation}
This gives $B_{\mathrm{new}}A_{\mathrm{new}}\approx\Delta W_{\mathrm{merge}}$ while preserving the standard LoRA loading interface. Because the five rank-8 experts span at most 40 dimensions per layer, the rank-40 recompression reproduces $\Delta W_{\mathrm{merge}}$ up to numerical error. On CogVideoX-2B, five rank-128 experts are merged exactly into a rank-640 adapter.

\section{Merge retention evaluation}
\label{app:retention}

To separate group training quality from merge-induced compression, we run a retention evaluation on the five group-specific LoRA experts. Each group-only LoRA is evaluated only on the VBench2.0 dimensions assigned to that group, and the resulting dimension scores are compared with the base model and the merged rank-40 adapter under a matched quick-evaluation protocol. This is an oracle-style diagnostic rather than a deployable single-model result: it asks whether the merged adapter preserves the capabilities learned by the group experts, rather than only whether the final merged model improves the average score.

\begin{table}[h]
\centering
\caption{Group-level retention summary. Group-only LoRAs are an oracle-style diagnostic evaluated only on assigned dimensions; Merged r40 (Ours) preserves the average while still compressing some group-specific skills. Bold indicates the highest score in each row.}
\begin{tabular}{lrrrrr}
\toprule
Scope & Base & Merged r40 (Ours) & Group-only & Merged -- Group & +OPD (Ours) \\
\midrule
All 17 dims & 47.07 & 48.18 & 48.06 & +0.12 & \textbf{49.91} \\
Group 1 & 60.03 & 62.98 & 57.11 & +5.87 & \textbf{64.11} \\
Group 2 & 66.39 & 64.89 & 63.10 & +1.79 & \textbf{66.41} \\
Group 3 & 26.70 & 27.40 & \textbf{27.47} & -0.07 & 27.37 \\
Group 4 & 36.86 & 38.80 & \textbf{42.72} & -3.92 & 42.10 \\
Group 5 & 33.46 & 35.09 & \textbf{41.46} & -6.37 & 37.98 \\
\bottomrule
\end{tabular}
\label{tab:retention_summary}
\end{table}

Table~\ref{tab:retention_summary} shows that the rank-40 merge preserves most average group-only performance but compresses some specialized skills. Groups 1 and 2 are better after merging than in their group-only form, suggesting that some dimensions benefit from updates learned by other groups. In contrast, Groups 4 and 5 lose 3.92 and 6.37 points after merging. The largest dimension-level compression occurs on multi-view consistency and mechanics, as shown in Table~\ref{tab:retention_dims}.

\begin{table}[h]
\centering
\caption{Selected dimension-level retention results from the group-only LoRA evaluation. Scores are percentages under the quick VBench2.0 protocol.}
\resizebox{\linewidth}{!}{%
\begin{tabular}{llrrrrr}
\toprule
Group & Dimension & Base & Merged r40 (Ours) & Group-only & Merged -- Group & +OPD (Ours) \\
\midrule
Group 1 & Dynamic Spatial Relationship & 34.00 & 38.00 & 32.00 & +6.00 & 38.00 \\
Group 1 & Human Clothes & 97.92 & 95.92 & 92.00 & +3.92 & 98.40 \\
Group 1 & Instance Preservation & 84.21 & 86.00 & 82.00 & +4.00 & 86.00 \\
Group 1 & Motion Order Understanding & 24.00 & 32.00 & 22.45 & +9.55 & 34.04 \\
Group 2 & Human Anatomy & 88.21 & 85.99 & 83.88 & +2.11 & 85.94 \\
Group 2 & Human Identity & 73.37 & 71.59 & 68.54 & +3.05 & 71.49 \\
Group 2 & Human Interaction & 66.00 & 62.00 & 58.00 & +4.00 & 68.00 \\
Group 2 & Motion Rationality & 38.00 & 40.00 & 42.00 & -2.00 & 40.20 \\
Group 3 & Camera Motion & 12.00 & 12.00 & 6.00 & +6.00 & 12.00 \\
Group 3 & Complex Landscape & 20.00 & 22.40 & 22.80 & -0.40 & 22.80 \\
Group 3 & Composition & 48.10 & 47.80 & 53.60 & -5.80 & 47.30 \\
Group 4 & Dynamic Attribute & 20.00 & 20.00 & 12.00 & +8.00 & 24.00 \\
Group 4 & Material & 36.67 & 32.14 & 30.00 & +2.14 & 32.70 \\
Group 4 & Multi-view Consistency & 42.96 & 51.95 & 73.56 & -21.60 & 60.02 \\
Group 4 & Thermotics & 47.83 & 51.11 & 55.32 & -4.21 & 51.69 \\
Group 5 & Complex Plot & 10.40 & 12.40 & 10.00 & +2.40 & 17.47 \\
Group 5 & Mechanics & 56.52 & 57.78 & 72.92 & -15.14 & 58.49 \\
\midrule
All & Mean & 47.07 & 48.18 & 48.06 & +0.12 & 49.91 \\
\bottomrule
\end{tabular}%
}
\label{tab:retention_dims}
\end{table}

Selecting post hoc, for each dimension, the better of the assigned expert and the merged adapter yields a diagnostic mean of 51.07. Table~\ref{tab:merge_diag} rules out rank truncation and broad cancellation of expert deltas as the main causes of specialist compression. Group 5 (complex plot and mechanics) has a small learned delta and a 2/17 merge weight, consistent with dilution of the mechanics specialist. Multi-view consistency belongs to Group 4, whose norm share is not small, so its compression is more consistent with averaging or activation-level interference.

\begin{table}[h]
\centering
\caption{Parameter-space merge diagnostics.}
\label{tab:merge_diag}
\begin{tabular}{lr}
\toprule
Diagnostic & Value \\
\midrule
Rank-40 SVD reconstruction error (Wan2.1) & $1.464\times10^{-6}$ \\
Pairwise cosine between global expert deltas & 0.730--0.938 \\
Modules with a negative expert-pair cosine & 0/240 \\
Group-size-weighted norm retention & 0.946 \\
Group 5 share of the weighted expert-delta norm & 2.08\% \\
CogVideoX-2B rank-640 merge error (dense delta / output) & $3.44\times10^{-7}$ / $7.05\times10^{-7}$ \\
\bottomrule
\end{tabular}
\end{table}

\section{On-policy distillation details}
\label{app:opd}

Each OPD step processes one training prompt per GPU. (i) The current student samples an 8-step trajectory from fresh Gaussian noise without gradient, and two interior latent states $z_\tau$ are stored with their timesteps. (ii) The prompt's dimension selects its group expert $\psi_{m(p)}$; the frozen expert and the frozen merged anchor $\phi_0$ predict velocities at the stored states. (iii) The student predicts velocities at the same states, and Eq.~\ref{eq:opd} is minimized with respect to the student LoRA only. All LoRA states share the same frozen base model and are swapped in place, so no additional backbone copy is required. The loss does not use the training videos, the flow-matching target of Eq.~\ref{eq:reward_loss}, or the reward weights. Table~\ref{tab:opd_params} lists the hyperparameters, and Table~\ref{tab:opd_delta} reports the change of every dimension relative to the merged adapter. The paired 95\% bootstrap interval of the mean change is $[+0.42, +3.05]$.

\begin{table}[h]
\centering
\caption{Change after OPD relative to the merged rank-40 adapter on Wan2.1.}
\label{tab:opd_delta}
\small
\begin{tabular}{lr@{\hspace{2.5em}}lr}
\toprule
Dimension & $\Delta$ & Dimension & $\Delta$ \\
\midrule
Multi-view Consistency & $+8.07$ & Thermotics & $+0.58$ \\
Human Interaction & $+6.00$ & Material & $+0.56$ \\
Complex Plot & $+5.07$ & Complex Landscape & $+0.40$ \\
Dynamic Attribute & $+4.00$ & Motion Rationality & $+0.20$ \\
Human Clothes & $+2.48$ & Dynamic Spatial Relationship & $0.00$ \\
Motion Order Understanding & $+2.04$ & Camera Motion & $0.00$ \\
Mechanics & $+0.71$ & Instance Preservation & $0.00$ \\
Human Anatomy & $-0.05$ & Human Identity & $-0.10$ \\
Composition & $-0.50$ & \textbf{Mean} & $\mathbf{+1.73}$ \\
\bottomrule
\end{tabular}
\end{table}

Relative to the six specialist gaps of the merged adapter, OPD recovers 8.07 of 21.61 points on multi-view consistency, 0.71 of 15.14 on mechanics, 0.58 of 4.21 on thermotics, 0.20 of 2.00 on motion rationality, and all 0.40 on complex landscape, while composition moves 0.50 further from its expert.

\section{CogVideoX-2B experiment}
\label{app:cogvideox}

The CogVideoX-2B experiment reruns the full procedure on a second backbone with its native diffusion-denoising objective: self-generated training videos, QA rewards, gradient probes, grouping, expert training, and merging. Joint and grouped training use the same 9,520 videos, prompts, reward weights, and LoRA and optimizer configuration (LoRA rank 128, $\alpha=128$, learning rate $1\times10^{-4}$); each sample is seen exactly once, because the five experts train on disjoint subsets whose sizes sum to 9,520. The five rank-128 experts are merged exactly into a rank-640 adapter ($\alpha=640$). Scores are VBench2.0 macro means over 17 dimensions; multi-view consistency is 0 for all CogVideoX methods.

\begin{table}[h]
\centering
\caption{CogVideoX-2B Joint LoRA across retained checkpoints (VBench2.0 macro score, $\times100$).}
\label{tab:cog_traj}
\begin{tabular}{lrr}
\toprule
Model / checkpoint & Training exposure & Score \\
\midrule
Base & 0\% & 39.87 \\
Joint LoRA & 25\% & 36.78 \\
Joint LoRA & 50\% & 31.58 \\
Joint LoRA & 75\% & 32.12 \\
Joint LoRA & 100\% & 27.95 \\
Grouped LoRA (Ours) & 100\% & 40.99 \\
\bottomrule
\end{tabular}
\end{table}

Joint LoRA is already below the base model at 25\% exposure, and even its best retained checkpoint remains below both the base model and the grouped adapter, so early stopping would not change the comparison. The degradation is dimension-selective: human anatomy (48.5 to 98.4), human identity (70.9 to 97.9), and thermotics (46.8 to 56.0) improve under Joint LoRA, while composition, camera motion, mechanics, human interaction, motion order, and motion rationality deteriorate. The grouped--Joint difference of $+13.04$ points has a paired prompt-cluster 95\% bootstrap interval of $[+9.08, +16.92]$.

\section{Independent evaluation with VideoScore}
\label{app:videoscore}

\begin{table}[h]
\centering
\caption{VideoScore-v1.1 on the matched Wan2.1 evaluation videos. The derived mean averages the five aspects; its paired 95\% CI is $[0.140, 0.192]$ with $P(\Delta>0)=1.000$.}
\label{tab:videoscore}
\begin{tabular}{lrrr}
\toprule
Aspect & Base & Grouped LoRA (Ours) & $\Delta$ \\
\midrule
Visual quality & 3.029 & 3.196 & $+0.168$ \\
Temporal consistency & 2.748 & 2.931 & $+0.183$ \\
Dynamic degree & 3.225 & 3.380 & $+0.155$ \\
Text alignment & 2.911 & 3.041 & $+0.130$ \\
Factual consistency & 2.659 & 2.851 & $+0.192$ \\
\midrule
Five-aspect mean & 2.914 & 3.080 & $+0.166$ \\
\bottomrule
\end{tabular}
\end{table}

\paragraph{Bootstrap protocol.} For every paired comparison, prompts are resampled with replacement as clusters, both videos of a prompt stay together, and the same prompt indices are used for both methods. We report the point estimate, the 95\% percentile interval over 10,000 resamples, and the fraction of resamples with a positive difference. For Grouped LoRA (Ours) versus Base on VBench2.0 this gives $+1.11$ points (bootstrap mean $+1.09$), a 95\% interval of $[-0.08, +2.26]$, and $P(\Delta>0)=0.966$.


\end{document}